\documentclass[sigconf]{acmart}

\usepackage{multirow}
\usepackage{subfig}
\usepackage[title]{appendix}
\usepackage{color}
\usepackage[utf8]{inputenc}
\usepackage{amsmath}
\usepackage{algpseudocode}
\usepackage{algorithm}
\usepackage{paralist}
\usepackage{makecell}
\usepackage[symbol]{footmisc}
\usepackage{multirow}
\usepackage{tabularx}
\usepackage{ulem}

\AtBeginDocument{%
  \providecommand\BibTeX{{%
    \normalfont B\kern-0.5em{\scshape i\kern-0.25em b}\kern-0.8em\TeX}}}

\setcopyright{acmcopyright}
\copyrightyear{2021}
\acmYear{2021}

\acmDOI{}
\acmConference[BIOKDD'21]{Singapore '21: 20th International Workshop on Data Mining in Bioinformatics}{August 15,2021}{Virtual Event, Singapore}
\acmBooktitle{Singapore '21: 20th International Workshop on Data Mining in Bioinformatics,
  August 15, 2021, Singapore}
\acmPrice{}
\acmISBN{}

\acmSubmissionID{7}

\begin{document}

\title{Multiple Organ Failure Prediction with Classifier-Guided Generative Adversarial Imputation Networks}

\author{Xinlu Zhang*}
\thanks{*Equal Contributors}
\affiliation{%
  \institution{Department of Computer Science, University of California, Santa Barbara}
  \city{Santa Barbara}
  \state{USA}
}
\email{xinluzhang@cs.ucsb.edu}

\author{Yun Zhao*}
\affiliation{%
  \institution{Department of Computer Science, University of California, Santa Barbara}
  \city{Santa Barbara}
  \state{USA}
}
\email{yunzhao@cs.ucsb.edu}

\author{Rachael Callcut}
\affiliation{%
  \institution{UC, Davis Health}
  \city{Davis}
  \state{USA}
}
\email{racallcut@ucdavis.edu}

\author{Linda Petzold}
\affiliation{%
  \institution{Department of Computer Science, University of California, Santa Barbara}
  \city{Santa Barbara}
  \state{USA}
}
\email{petzold@cs.ucsb.edu}

\renewcommand{\shortauthors}{Xinlu Zhang, Yun Zhao, Rachael Callcut and Linda Petzold }
\begin{abstract}



Multiple organ failure (MOF) is a severe syndrome with a high mortality rate among Intensive Care Unit (ICU) patients. Early and precise detection is critical for clinicians to make timely decisions. An essential challenge in applying machine learning models to electronic health records (EHRs) is the pervasiveness of missing values. Most existing imputation methods are involved in the data preprocessing phase, failing to capture the relationship between data and outcome for downstream predictions. In this paper, we propose classifier-guided generative adversarial imputation networks (Classifier-GAIN) for MOF prediction to bridge this gap, by incorporating both observed data and label information. Specifically, the classifier takes imputed values from the generator(imputer) to predict task outcomes and provides additional supervision signals to the generator by joint training. The classifier-guide generator imputes missing values with label-awareness during training, improving the classifier’s performance during inference. We conduct extensive experiments showing that our approach consistently outperforms classical and state-of-art neural baselines across a range of missing data scenarios and evaluation metrics.
\end{abstract}




\maketitle

\section{Introduction}
Multiple organ failure (MOF) is a life-threatening syndrome with variable causes, including sepsis~\cite{rossaint2015pathogenesis},  pathogens~\cite{harjola2017organ}, and complicated pathogenesis~\cite{wang2018clinical}. It is a major cause of death in the surgical intensive care unit (ICU)~\cite{durham2003multiple}. Care in the first few hours after admission is critical to patient outcomes. This period is also more prone to medical decision errors in ICUs than later times~\cite{ICU}. Thus, an effective and real-time prediction is essential for clinicians to provide appropriate treatment and increase the survival rates for MOF patients. 
\\

\begin{figure}[ht]
  \includegraphics[width=\linewidth]{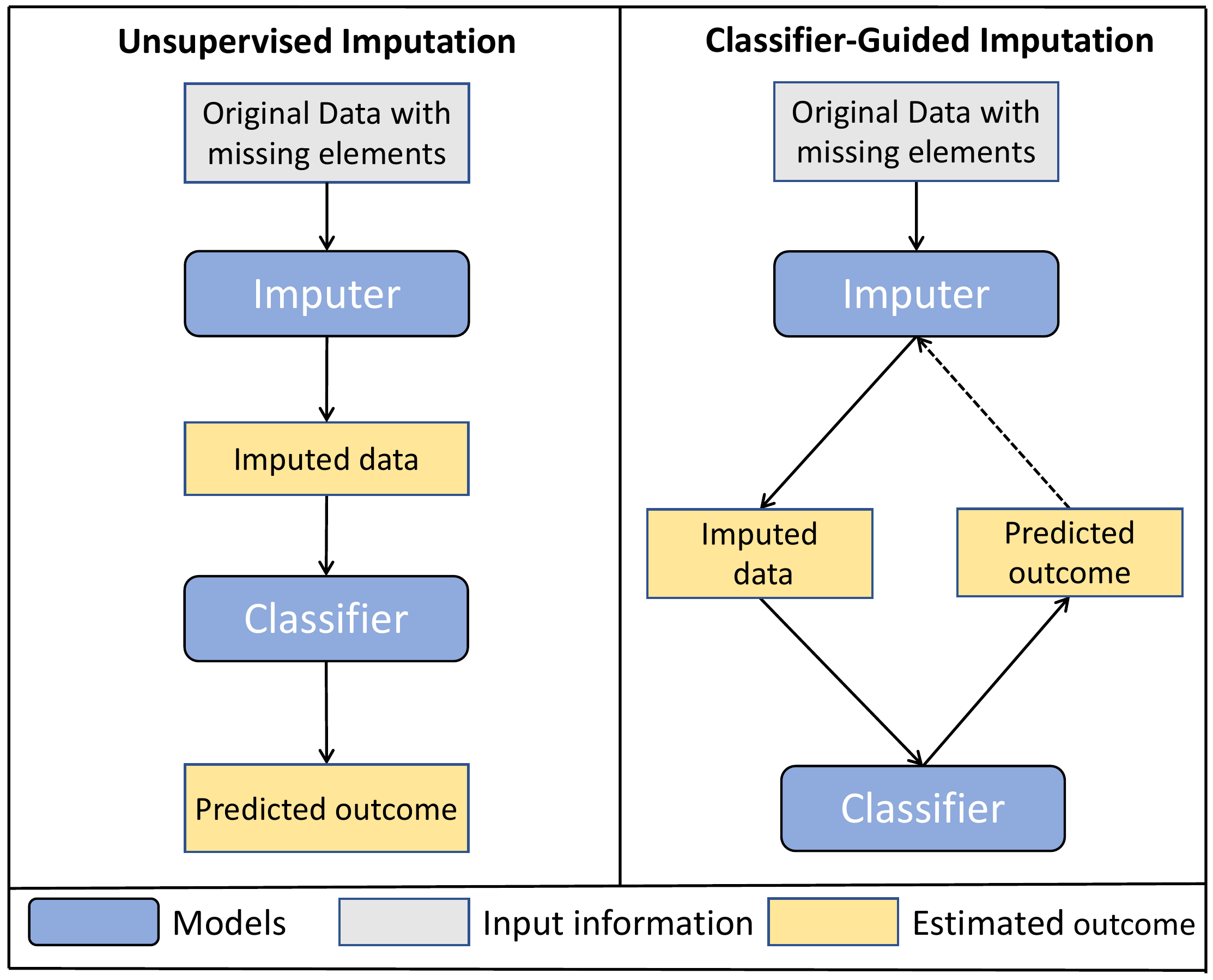}
  \caption{ Workflow comparison of unsupervised imputers (left) and Classifier-GAIN (right). Classifier-guided imputers learn from the classifier during training and improve classification during inference, while unsupervised imputers learn only from the partially observed data in the data-prepossessing phase. The solid lines represent processes that occur during both the training and inference phase, while the dashed line represents the step that only occurs during training. Note that our final goal is to improve the classifier's performance utilizing imputed data.}
  \label{outline}
\end{figure}

With the rapid growth of electronic health record (EHR) data availability, machine learning models have drawn increasing attention for MOF prediction. Missing values are a pervasive and serious medical data issue, which could be caused by various reasons such as lost records or inability to collect the data during some time periods
~\cite{zhang2016identification}. 
There exist many imputation algorithms, such as mean value imputation~\cite{acuna2004treatment}, multivariate imputation by chained equations (MICE)~\cite{MICE} and generative adversarial imputation nets (GAIN)~\cite{yoon2018gain} which impute missing components by adapting generative adversarial networks (GANs). 
However, these methods focus only on constructing the distribution between the unobserved components and the observed ones, without considering the underlying connections with specific downstream tasks, as is shown in the Figure~\ref{outline}.\\
\vspace*{-\baselineskip}

Recently, GANs~\cite{goodfellow2014generative} have made significant progress on data generation. Labels can be incorporated in the GAN framework, e.g. CGAN~\cite{mirza2014conditional} and AC-GAN~\cite{odena2017conditional}, to generate label-aware outputs.
 Semi-supervised GAN ~\cite{odena2016semi} introduces a classifying discriminator to output either the validity of data or its class.
Triple-GAN ~\cite{li2017triple} is further proposed by adding an additional classifier to separate the role of the discriminator in Semi-supervised GAN. These works leverage label information to improve data generation and their generators have to take ground truth labels to generate label-aware data, which is not applicable in classification problems during inference.


In this paper, we propose a classifier-guided missing value imputation framework for MOF prediction with early-stage measurements after admission. The generator uses observed data and random noise to impute missing components and obtains imputed data; the classifier takes the generator’s outcomes, models the relationship between imputed data and labels by joint training with the generator, and outputs estimated labels. The discriminator attempts to identify which component is observed by taking imputed data from the generator and predicted label information from the classifier.\\


The key contributions of this paper include:

1) We propose a classifier-guided missing value imputation deep learning framework for MOF prediction, which incorporates both observed data and label for modeling label-aware imputation during training to help classification during inference. To the best of our knowledge, this is the first GAN-based end-to-end deep learning architecture for MOF prediction with missing values.


2) Experimental results on both synthetic and real-world MOF datasets show that our Classifier-GAIN outperforms GAIN and MICE consistently in different missing ratio scenarios and evaluation metrics.

3) Visualization of the values imputed by our approach further validates the effectiveness of Classifier-GAIN compared to various baselines.

The remainder of our paper is organized as follows. Preliminaries are introduced in Section~\ref{background}, followed by the details of our proposed
approach in Section~\ref{method}. Experimental results are reported in Section~\ref{Results}. In Section~\ref{related}, we review the existing related work, and conclusions are given in  Section~\ref{discussion}.



\section{Preliminaries}\label{background}
We formulate the MOF prediction as a binary classification problem with missing components in multiple features. In this section, we describe the problem definition in Section~\ref{probdefin} and review the GAIN imputation algorithm in Section~\ref{gain}. The related notations are summarized in Table~\ref{tab:notation}.  Specifically, throughout this paper, we utilize lower-case letters, e.g. $\textbf{x}$, to denote the data vector.
$p(\textbf{x})$ is the probability distribution function of $\textbf{x}$.
$\textbf{1}$ denotes a d-dimensional vector of $1$s, and letters with hats such as $\hat{\textbf{x}}$  denote estimated vectors.

\begin{table}[ht]
  \caption{Notation definitions}
  \label{tab:Notations}
  \setlength{\tabcolsep}{.1pt}
  \begin{tabular}{ccl}
    \toprule
    Notations&Description\\
    \midrule

    $i$ &  index of observations\\
     $j$ &  index of observed features\\
     $d \in \mathbb{N}$ & number of  observed features\\
    $N \in \mathbb{N}$ & total number of observations\\
      $n \in \mathbb{N}$ & size of minibatch\\
    $\textbf{x}$ &  data vector \\ 
    $y$ & outcome indicator \\
    $\textbf{m}$ & mask vector\\
    $\textbf{z}$ &  noise vector  \\
    $\textbf{h}$ &  hint vector \\
    $\textbf{b}$ &  binary vector for calculating hint \\
     $\tilde{\textbf{x} }$ & combination of partially observed data and \textit{NA}s \\
       $\ddot{\textbf{x} }$ & combination of partially observed data and noise\\
       $G$ & generator\\
      $C$ & classifier \\
     $D$ & discriminator\\
     
     $\textbf{g} $ & reconstructed vector, the output of \textit{G}\\ 
    $\hat{\textbf{x}}$ & imputed data vector \\
    $\hat{\textbf{m}}$ & estimated mask, the output of \textit{D} \\
     $\hat{y}$ & estimated label, the output of \textit{C}  \\

  \bottomrule
\end{tabular}
\label{tab:notation}
\end{table}
\vspace*{-\baselineskip}

\subsection{Problem definition} \label{probdefin}

Let $\mathcal{X}^d$ be a $d$-dimensional space  and $\textbf{x}$ a data vector, taking values in $\mathcal{X}^d$ following distribution $p(\textbf{x})$. We denote $x_{j}$ as the $j$-th feature in $\textbf{x}$. Binary mask vector $\textbf{m} \in  \{0,1\}^d$ indicates if the corresponding element in $\textbf{x}$ is missing or not, where $x_j$ is observed if  $m_{j}=1$, otherwise  $x_{j}$ is missing.  To clarify the observed and missing components, we define a new vector $\tilde{\textbf{x}}=(\tilde{x}_1, \cdots, \tilde{x}_d)$ as follows:

\[
    \tilde{x}_{j\in \{1, 2, ..., d\}}= 
\begin{cases}
    x_{j},& \text{if }  m_j =1, \\
     \textit{NA},              & \text{if }  m_j=0.
\end{cases}
\]
Supposing that $y \in \{0,1\}$ is the binary outcome indicator for each sample, we can represent the dataset as a collection of $N$ i.i.d. samples $ \{(\tilde{\textbf{x}}_i,\textbf{m}_i),y_i\}_{i=1}^N$.

We aim to impute the missing components in every $\tilde{\textbf{x}}$,
and predict the outcome $y$ for all samples by leveraging the imputed data vector. 
Formally, we seek to model $p(y|\tilde{\textbf{x}})$: the conditional distribution of the task outcome given a partially observed data vector.

\subsection{ Generative adversarial imputation networks (GAIN)}\label{gain}

GAIN \cite{yoon2018gain} was proposed to impute missing components with a GAN framework. In GAIN, the generator takes the observed components in $\tilde{\textbf{x}}$, mask vector $\textbf{m}$ and a noise vector $\textbf{z}$  as inputs, and outputs a completed data vector. The discriminator tries to distinguish the observed components and the missing ones. Furthermore, a hint vector $\textbf{h}$ is introduced to provide additional missing information for alleviating the diversity of the imputation result. 

The generator, \textit{G},  takes

$$\ddot{\textbf{x}}=\textbf{m} \odot  \tilde{\textbf{x}}+ (\textbf{1}-\textbf{m}) \odot \textbf{z},$$
a combination of  $\tilde{\textbf{x}}$  and  \textbf{z} by element-wise multiplication with  $\textbf{m}$, as input, and outputs \textbf{g}, the reconstructed vector,
$$\textbf{g}=G(\ddot{\textbf{x}}).$$
Note that $\textbf{g}$ is an output vector for every component, even if values are not missing in the data vector.

Thus, another element-wise multiplication is performed to calculate the imputed data vector via 
$$\hat{\textbf{x}} =\textbf{m} \odot  \tilde{\textbf{x}}  +  (\textbf{1}-\textbf{m})\odot \textbf{g}, $$
where $\hat{\textbf{x}}$ is obtained by taking the observed part in $\tilde{\textbf{x}}$ and replacing each \textit{NA} by the corresponding value in $\textbf{g}$.

The discriminator serves as an adversarial character to train \textit{G} by taking in the imputed data vector $\hat{\textbf{x}}$ and the hint vector $\textbf{h}$, following the distribution $p(\textbf{h}| \textbf{m})$. The output of the discriminator is a distribution to identify which components in $\hat{\textbf{x}}$ are observed. To help the discriminator distinguish imputations and observations, $\textbf{h}$ provides certain information about $\textbf{m}$ and its amount can be controlled by adjusting $\textbf{h}$ in different settings. Specifically, 
a binary random variable $ \textbf{b}\in \{0,1\}^d$ is randomly drawn with  $P(b_{j}=1) =p$. Then,  $\textbf{h}|\textbf{m}$  is calculated by
 $$\textbf{h}  = \textbf{m}  \odot  \textbf{b} + 0.5 (\textbf{1}-\textbf{b}),$$ such that the discriminator will get mask information by  $h_{j}=m_{j}$ if $b_{j}=1$, otherwise, no information provided.

\section{Methodology}\label{method}
\begin{figure}[ht]
  \includegraphics[width=\linewidth]{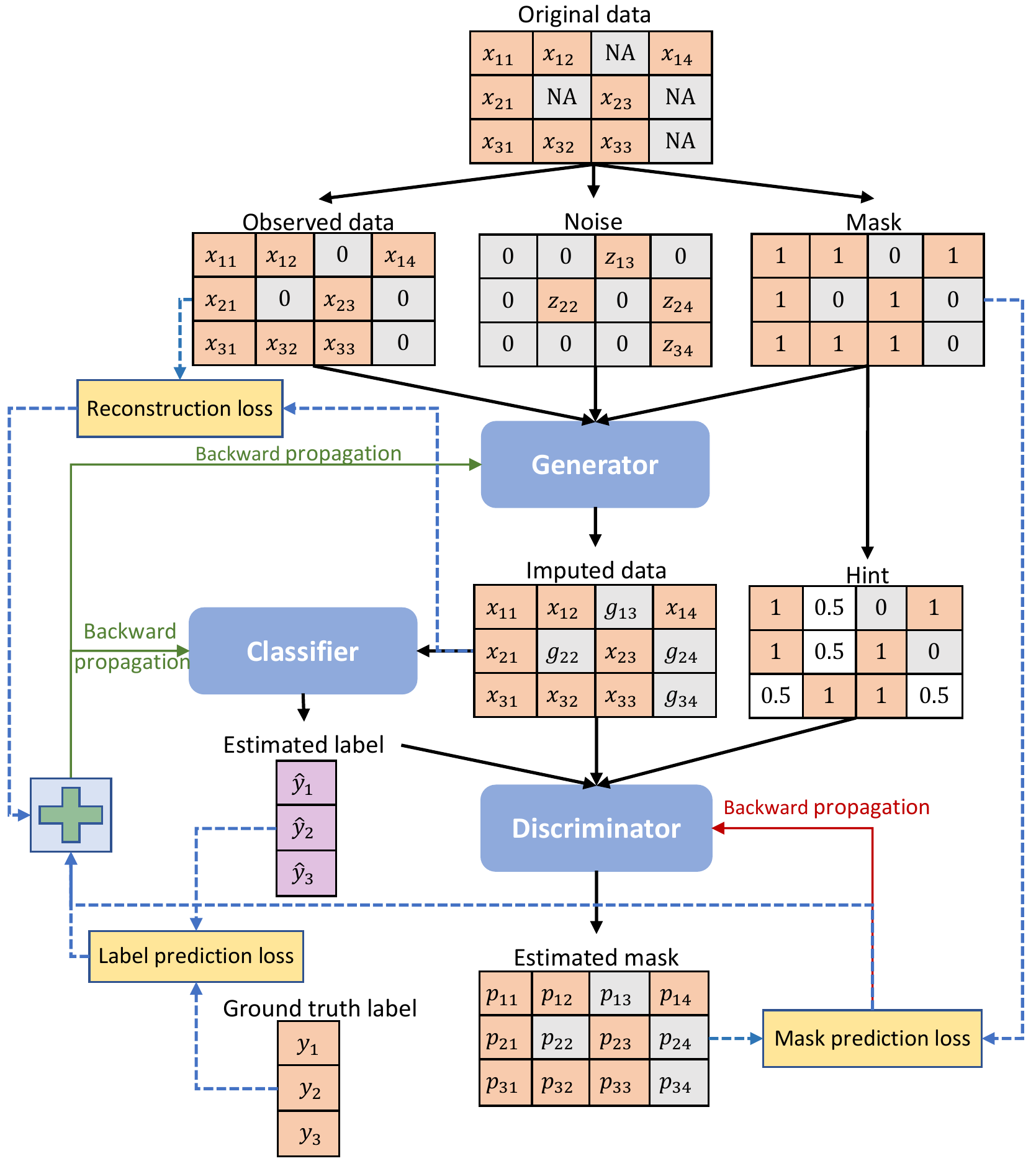}
  \caption{The overall architecture of Classifier-GAIN with three samples (adapted from GAIN  \cite{yoon2018gain}).
Each row in the matrices and vectors corresponds to a sample. The input of our network is the original data vector with missing components. The generator takes partial observation data, a noise matrix filling missing components and mask, and outputs imputed data, which is then fed into the classifier to obtain the estimated labels. The imputed data and the estimated labels along with a hint matrix are fed into the discriminator to estimate the mask. Dashed lines represent information flows for loss calculation. Green solid lines represent backward propagation for the generator and the classifier training, and the red solid line represents the backward propagation for the discriminator training.}
  \label{architecture}
\end{figure}

Notably, conditional information such as labels can improve the performance of the generator~\cite{mirza2014conditional, odena2017conditional}, and the completed data vector can enhance its task prediction result. However, state-of-art imputation methods, e.g. GAIN, do not make use of the relationship between observations and outcome labels, which could provide additional information to help downstream classification tasks. Therefore, we propose Classifier-GAIN to bridge this gap. Figure~\ref{architecture} depicts the overall architecture.
We explain each of the components and the training process of Classifier-GAIN in detail in Section~\ref{method:cgain}$\sim$~\ref{method:train}. 

\subsection{Classifier-guided generative adversarial imputation nets} \label{method:cgain}
In an imputation setting, we propose to fill in the missing components $\textit{NA}$s in $\tilde{\textbf{x}}$, using the distribution of data obtained by the generator, \textit{G}. To guide the data imputation and predict the final task outcome, the classifier, \textit{C}, takes imputed data and is trained together with \textit{G}.
The discriminator \textit{D} plays an adversarial role to train G, with additional label prediction information from \textit{C}.

\subsubsection*{\textbf{Generator and Classifier}}\label{method:g}

Similar to the structure of GAIN, \textit{G} takes $\tilde{\textbf{x}}$ as input and outputs $\hat{\textbf{x}}$, trying to model $p(\textbf{x}|\tilde{\textbf{x}})$, the conditional distribution of a data vector given the partial observations. The classifier \textit{C} is a supervised learning model to predict the task outcome, $\hat{y}$, by taking $\hat{\textbf{x}}$ from \textit{G}, which obtains the conditional distribution $P(y|\textbf{x})=P(y|\hat{\textbf{x}})$.
We define the estimated outcome $\hat{y}$ by
$$\hat{y} = C(\hat{\textbf{x}}).$$
Then \textit{G} and \textit{C} are jointly trained to obtain the distribution $p(y,\textbf{x} |\tilde{\textbf{x}})$, making \textit{G} label-aware during imputation, which is ignored by GAIN.

\subsubsection*{\textbf{Discriminator}}\label{method:d}
In our architecture, the discriminator \textit{D} serves as an adversarial character to train \textit{G} by receiving the predicted label information from  \textit{C}. We input $ \hat{\textbf{x}}$, $\hat{y}$ and $\textbf{h}$ into \textit{D} to obtain the probability that each component in $\hat{\textbf{x}}$ is observed. Here, $ \hat{\textbf{x}}$ and $\hat{y}$ jointly provide information to enhance \textit{D} by learning the relationship between data and task outcomes, which can further strengthen \textit{G} and \textit{C}. We define the estimated mask variable, $\hat{\textbf{m}} \in [0,1]^d$, by$$\hat{\textbf{m}}= D (\hat{\textbf{x}}, \hat{y}, \textbf{h}),$$
with the $j$-th item in $\hat{\textbf{m}}$ corresponding to the probability that the $j$-th item in $\hat{\textbf{x}}$ is not \textit{NA} in $\tilde{\textbf{x}}$.


\subsection{Classifier-GAIN training }\label{method:train}
\textit{G}, \textit{C} and \textit{D} are trained as a min-max game by
\begin{multline}
\min_{G,C} \max_{D} V(G,C,D) =\mathbb{E}_{\ddot{\textbf{x}}, \textbf{m}, \textbf{h}}
\Bigg[\textbf{m} \log \Big(D \big( G(\ddot{\textbf{x}}), C(G(\ddot{\textbf{x}})), \textbf{h}\big)\Big)   \\ +(\textbf{1}-\textbf{m} ) \log \Big(\textbf{1}- D \big(G(\ddot{\textbf{x}}), C(G(\ddot{\textbf{x}})), \textbf{h} \big) \Big)
\Bigg],
\end{multline}
where $\log$ is an element-wise logarithm.  Specifically, We train  \textit{G} and \textit{C} together to minimize the probability of \textit{D} identifying $\textbf{m}$, 
maximize the probability of correctly predicting $y$ and minimize the reconstruction loss of observed components. We train \textit{D} to maximize the probability of correctly predicting $\textbf{m}$.





On each iteration,
\textit{G} and \textit{C} are updated $k$ times with objective function, $\mathcal{L}_{C\&G}$, which is a  weighted sum of three losses
\begin{equation}
\mathcal{L}_{C\&G}=\mathcal{L}_{G}(\textbf{m},\hat{\textbf{m}})+\alpha\mathcal{L}_R(\textbf{x},\hat{\textbf{x}}) +\beta \mathcal{L}_{C}(y,\hat{y}),\label{eq:3}
\end{equation}
where  $\alpha$ and $\beta$ are  hyper-parameters. \\

The first loss, $\mathcal{L}_{G}$, is an adversarial loss, which applies to missing components ($m_{j}=0$) by
\begin{equation}
\mathcal{L}_{G}(\textbf{m},\hat{\textbf{m}})=-\sum_{j=1}^d \Big[ (1-m_{j}) \log( \hat{m}_{j})\Big].
\end{equation}

The second loss, $\mathcal{L}_R$, is a reconstruction loss, which applies to observed components ($m_{j}=1$) by

\begin{equation}
\mathcal{L}_R(\textbf{x},\hat{\textbf{x}}) = \sum_{j=1}^d m_{j} L_R(x_{j}, \hat{x}_{j}),
\end{equation}
where

\[
    L_R(x_{j}, \hat{x}_{j})= 
\begin{cases}
    (x_{j} - \hat{x}_{j})^2,& \text{for numerical variables,} \\
    -x_{j} \log(\hat{x}_{j}),              & \text{for binary variables. }
\end{cases}
\]

The third loss, $\mathcal{L}_{C}$, is a binary cross entropy loss for task prediction given by

\begin{equation}
\mathcal{L}_{C}(y,\hat{y})=-\big[y\log{(\hat{y})} +(1-y) \log (1-\hat{y} ) \big].
\end{equation}
Note that as  \textit{G} and \textit{C} are updated together via Eq.~\ref{eq:3}, \textit{C} 's performance will influence \textit{G}'s  parameters to guide the missing component imputation.

\textit{D} updates once at each iteration  with  objective function
\begin{equation}
\mathcal{L}_{D}(\textbf{m},\hat{\textbf{m}})=-\sum_{j=1}^d  \Big[m_{j} \log( \hat{m}_{j}) +(1- m_{j}) \log(1-\hat{m}_{j}) \Big]. \label{eq:2}
\end{equation}
The detailed training process is shown in Algorithm 1.



  \begin{algorithm} \label{alg1}
    \caption{Minibatch Classifier-GAIN training }
    \begin{algorithmic}[1]
   \State \textbf{Input:} Original data vector with missing component $\tilde{\textbf{x}}$, mask vector $\textbf{m}$, ground truth label $y$, the probability for drawing the hint vector $\textit{p}$. hyper-parameters $\alpha$ and $\beta$. 
\Repeat 

\State \textbf{Generator and Classifier}
 \State Sample a batch of \textit{n} binary vector $\{\textbf{b}_i\}_{i=1}^n\sim \text{Bern}(p) ^d$
  \For {k steps}
  \State  Sample a batch of \textit{n} noises $\{\textbf{z}_i\}_{i=1}^n  \sim U(0,1]^d$ 
\For {$ i \gets 1 \text{ to } n$}
    \State $\textbf{h}_i \gets \textbf{m}_i \odot  \textbf{b}_i  + 0.5 (\textbf{1}-\textbf{b}_i)$

    \State  $\ddot{\textbf{x}}_i \gets \textbf{m}_i  \odot  \tilde{\textbf{x}}_i  + (\textbf{1}-\textbf{m}_i ) \odot \textbf{z}_i $
    \State Imputed data $\textbf{g}_i \gets G(\ddot{\textbf{x}}_i)$ \State  $\hat{\textbf{x}}_i \gets  \textbf{m}_i \odot \tilde{\textbf{x}}_i + (\textbf{1}-\textbf{m}_i) \odot  \textbf{g}_i$
    \State Obtain  $\hat{y}_i  \gets  C(\hat{\textbf{x}}_i )$ 
    \State Obtain  $\hat{\textbf{m}}_i \gets D(\hat{\textbf{x}}_i,\hat{y}_i,\textbf{h}_i)$
    \EndFor
\State Update generator $G$ and classifier $C$ together via stochastic gradient descent (SGD):
       \State $\bigtriangledown_G~\frac{1}{n}\sum_{i=1}^n  \mathcal{L}_G(\textbf{m}_i, \hat{\textbf{m}}_i) + \alpha\mathcal{L}_R(\textbf{x}_i,\hat{\textbf{x}}_i)+\beta \mathcal{L}_{C}(y_i,\hat{y}_i)$

  \EndFor
   \State
\State \textbf{Discriminator}
\State Sample a batch of \textit{n} binary vector $\{\textbf{b}_i\}_{i=1}^n\sim \text{Bern}(p) ^d$
\For {$ i \gets 1 \text{ to } n$}
\State  
$\textbf{h}_i \gets \textbf{m}_i \odot \textbf{b}_i  + 0.5 (\textbf{1}-\textbf{b}_i)$
\State Obtain  $\hat{\textbf{m}}_i \gets D(\hat{\textbf{x}}_i,\hat{y}_i,\textbf{h}_i)$
\EndFor
\State Update discriminator with fixed \textit{G} and \textit{C} via SGD
        \State $\bigtriangledown_D~\frac{1}{n}\sum_{i=1}^n \mathcal{L}_D(\textbf{m}_i,\hat{\textbf{m}}_i) $    
  
 \Until{Classifier-GAIN  converges}
     \end{algorithmic}
    \end{algorithm}
 
 \section{Experiments}\label{Results}
In this section, we conduct experiments on two datasets: the PhysioNet sepsis synthetic dataset and the UCSF real-world EHR dataset, introduced in Section~\ref{Dataset}, to evaluate Classifier-GAIN's performance. 
Particularly, we investigate
\begin{enumerate}[Q1.]
    \item Does the classifier-guided imputation help the downstream MOF prediction?
    \item How does the proposed algorithm perform across different missing ratio scenarios?
\end{enumerate}

We explain the experimental settings in Section~\ref{experiment_set}. The performance comparisons of Classifier-GAIN against other imputation algorithms for MOF prediction are shown in Section~\ref{sec:pc}, followed by the visualizations of the imputed missing values of the UCSF MOF dataset in Section ~\ref{case}.

\subsection{Dataset}\label{Dataset}


For the PhysioNet sepsis dataset, 10,587 patients and 40 features are contained. We randomly select $80\%$ of the instances as the training set, $10\%$ as the development set, and $10\%$  as the testing set. For the UCSF MOF dataset, 2,160 patients and 29 features are contained. we perform a 5-fold cross validation, considering the dataset's size and models' training time. The detailed description of datasets as follows:

\subsubsection*{\textbf{PhysioNet sepsis synthetic dataset}}\label{Sepsis}
Sepsis is  a severe critical illness syndrome that can result in MOF~\cite{rossaint2015pathogenesis}. Since MOF is the fatal end of sepsis progression\cite{bravo2019machine}, early detection of sepsis and antibiotic prescription are critical for improving MOF patient outcomes. We built a synthetic dataset based on the \textit{physiological data}~\cite{reyna2019early} provided by PhysioNet, sourced from ICU patients. 
Each patient contains $40$ hourly measurements in three categories (vital signs, laboratory values and demographics) and the sepsis outcome in each hour. To obtain the sepsis outcome in the early stages, we focus on the first 6 hours' records of each feature. We take the first-appearance measurement of each feature in the first six hours after admission. If the value of a feature was not recorded in the first six hours, we assume that value was missing. We exclude the patients whose features were entirely missing in the first six hours. 
We label a patient with sepsis as 1, otherwise as 0. To obtain a completed synthetic dataset for further experiments, we apply KNN (with $K = 5$) to impute the original missing components and SMOTE to balance the data. After data preprocessing, we obtained 10,587 patients, among which 5,808 patients are with sepsis and 4,779 without sepsis.


\subsubsection*{\textbf{UCSF MOF real-world dataset}} \label{UCSF}
Our UCSF MOF dataset,  collected from the  UCSF/San  Francisco  General  Hospital and Trauma Center, contains 2,190 patients admitted to a Level  I trauma center. Both demographic information, such as gender, age, BMI (body mass index), and injury measurements, e.g. injury severity score (ISS), traumatic brain injury, and Glasgow Coma Scale (GCS),  were measured at the admission time of each patient.  Laboratory results (D-Dimer, creatinine, white blood cell etc.) and physical vital signs (for example heart rate, respiratory rate, systolic blood pressure etc.)  were recorded at different hours. Unique ICU treatments such as blood transfusion units, fresh frozen plasma transfusion and crystalloids for fluid resuscitation were slotted into time intervals such as 0 to 24 hours. Medical treatments (vasopressor, Heparin and Factor VII et al.) were reported daily after admission.

To analyze the MOF states associated with patients' early-stage status, we select either the first day or the initial hour records manually. We extract features with importance scores higher than 2\% using forests of trees in Scikit-learn~\cite{pedregosa2011scikit}, and remove the patients whose data were utterly missing in the early stage or whose MOF outcome was not recorded. After data preprocessing and removal of abnormal values, we are left with 2,160 patients and 29 measurements. Selected features are categorized by types, 
and detailed statistics are shown in Appendix A. Two blood test features, D-Dimer and Factor VII, had a missing rate higher than $40\%$. The body mass index (BMI) missed $17.4\%$. Factor VII treatment, partial thromboplastin time (PTT), respiratory rate and systolic blood pressure were missing at rates between $5\%$ and $10\%$.  The remainder of the features were missing at rates less than $5\%$. The rate of missing data for each feature is listed in Table~\ref{tab:Missing}, ordered from high to low. 
Missing values account for $6.42\%$ among all observations 
and the labelling ratio between No MOF (class 0) and MOF (class 1) is $11 : 1$ in the dataset.
\vspace*{-\baselineskip}
\begin{table}
  \caption{Rates of missing data in the UCSF dataset}
  \label{tab:Missing }
    \setlength{\tabcolsep}{.5pt}
  \begin{tabular}{c l}
    \toprule
  Missing rate &  Feature\\
    \midrule
      41.9$\%$ & D-Dimer\\
     41.6$\%$& Factor VII (blood test)\\
	  17.4$\%$ & BMI  \\ 
     $ 10\%\geq \&  > 5 \%  $ &  Factor VII treatment, PTT, Respiratory, SBP\\
     $ 5\% \geq  \&  > 0\% $ & HR, numribfxs, GCS, Vasopressor, Bun, Serumco2, \\ 
     &  PLTs, Crystalloids, Crystalloids, WBC,
     HGB, HCT, \\& AIS scores, FFP\_units, Blood\_units, age, iss,\\ &  Thromboembolic complication, Heparin\_gtt\\
     $0\%$ &  Gender\\
            \bottomrule
\end{tabular}
\label{tab:Missing}
\end{table}

\subsection{Experimental settings}\label{experiment_set}

\subsubsection*{\textbf{Evaluation metrics}} \label{eval_metric}
We measure the performance of Classifier-GAIN and baselines by 
 both macro F1-score and area under the ROC curve (AUC-ROC).

Macro F1-score  is defined as the mean of class-wise F1 scores which
assigns equal importance to every class. It is low for models that perform well on the common classes, while performing poorly on the rare classes. For the Macro F1-Score calculation, we use 0.5 as the predicted value threshold.

AUC-ROC is also commonly used for MOF prediction~\cite{bakker1996serial,papachristou2010comparison}. AUC-ROC assesses the overall preference of a classifier by summarizing over all possible classification thresholds. In binary classification problems, the higher the AUC-ROC, the better the model's performance in identifying the two classes.

\begin{table}
  \caption{Hidden layer setting of different modules in UCSF and Sepsis datasets.}
      \setlength{\tabcolsep}{.5pt}
  \label{tab:config }
  \begin{tabular}{c c c c c}
    \toprule
     Dataset & Network &  Hidden layer 1 &  Hidden layer 2 & Dropout rate \\
     \hline
   \multirow{3}{*}{UCSF}  & Classifier &32&16&0.1\\
  &Generator &64&32&0.1\\
   &Discriminator &64&32&0.1\\
 \hline
   \multirow{3}{*}{Sepsis}  & Classifier &128&64&0.1\\
  &Generator &64&32&0.1\\
   &Discriminator &64&32&0.1\\
     \bottomrule
   
\end{tabular}
\label{tab:config}
\end{table}

\begin{table*}[h]
\centering
\setlength{\tabcolsep}{.1pt}
  \begin{tabular}{ c c |c |c| c || c |c |c| c } 
 \hline
     \multirow{3}{*}{\makecell{Missing \\ Rate}}
     &  \multicolumn{4}{c|| }{ macro F1-score }  &  \multicolumn{4}{c}{ AUC-ROC}\\
   \cline{2-9}
  {} & Classifier-GAIN  &  \makecell{Simple \\ imputation} & MICE& GAIN & Classifier-GAIN  &  \makecell{Simple \\ imputation} & MICE& GAIN\\
  \hline
    $0\%$  & \multicolumn{4}{c||}{Upper bound: $84.8 \pm 0.1$} & \multicolumn{4}{c }{Upper bound: $90.9 \pm 0.3$}\\
      \hline
  $20\%$  & \textbf{83.2} $\pm$ \textbf{0.3} &67.8 $\pm$ 2.1 &68.6 $\pm$ 1.5 & 69.5 $\pm$ 1.8 & \textbf{88.3} $\pm$ \textbf{0.6}&83.1 $\pm$ 0.9 &84.3 $\pm$ 1.5 &83.4 $\pm$ 1.4\\
  $25\%$ &\textbf{82.9} $\pm$ \textbf{0.2} &68.3 $\pm$ 3.3& 66.9 $\pm$ 1.5 &65.9 $\pm$ 1.6  & \textbf{87.1} $\pm$ \textbf{0.4}  &81.8 $\pm$ 1.2&81.9 $\pm$ 0.6 &81.3 $\pm$ 0.6  \\
  $30\%$  &\textbf{80.8} $\pm$ \textbf{0.8} &67.4 $\pm$ 2.1 & 67.3 $\pm$ 1.1 &67.1 $\pm$ 2.0 &\textbf{86.4} $\pm$ \textbf{0.6} &81.9 $\pm$ 1.0& 81.0 $\pm$ 0.9 &81.8 $\pm$ 1.0 \\
  $35\%$  &\textbf{79.7} $\pm$ \textbf{0.9}&66.9 $\pm$ 1.6& 68.9 $\pm$ 1.5& 66.2 $\pm$ 3.3 &\textbf{84.8} $\pm$ \textbf{0.4}&81.2 $\pm$ 1.1&80.3 $\pm$ 1.0 & 81.7 $\pm$ 0.8  \\
  $40\%$ &\textbf{78.8} $\pm$ \textbf{0.7} &64.6 $\pm$ 2.6&  66.9 $\pm$ 2.3 &64.8 $\pm$ 3.5 &\textbf{83.9} $\pm$ \textbf{0.6} &80.3 $\pm$ 0.4&  80.0 $\pm$ 1.1 &81.2 $\pm$ 0.9 \\
  $45\%$ &\textbf{77.2} $\pm$ \textbf{0.5}  &65.6 $\pm$ 3.0&  66.4 $\pm$ 1.9 &65.4 $\pm$ 3.2 & \textbf{81.1} $\pm$ \textbf{0.4} &79.5 $\pm$ 0.9& 79.2 $\pm$ 0.9 &79.0 $\pm$ 0.7 \\
  $50\%$  & \textbf{77.7} $\pm$ \textbf{0.8}&65.1 $\pm$ 1.6 &  63.7 $\pm$ 1.0 &65.5 $\pm$ 2.1  &  \textbf{82.1} $\pm$ \textbf{0.7} &80.5 $\pm$ 0.4& 74.7 $\pm$ 0.6  &80.7 $\pm$ 0.6 \\
\hline
     \bottomrule
\end{tabular}

\caption{Model performance (\%) on PhysioNet sepsis dataset in different missing ratio settings.}
\label{tab:sepsisprof}
\end{table*} 
 
\subsubsection*{\textbf{Model configurations}} \label{model_config}
We compare Classifier-GAIN with both classical and state-of-art neural baselines for MOF prediction as follows:

\begin{enumerate}

\item \textbf{Simple imputation}:
It imputes missing components by mean imputation and most frequent imputation for continuous and  categorical variables, respectively.

\item \textbf{MICE}~\cite{MICE}: It is a multiple imputation method, accounting for the statistical uncertainty in the imputations.

\item \textbf{GAIN} \cite{yoon2018gain}: It is a deep learning adversarial imputation framework, which we explained in Section \ref{gain} in detail.
\end{enumerate}
Each of methods (1), (2) and (3) is separated into two steps. First we impute missing components by the corresponding method. Then we utilize a binary classifier to predict the subjects' outcomes by taking imputed data.  For our proposed Classifier-GAIN, we take the partially observed data as input, and output both an imputed data and classification outcomes.

In order to make the performance comparison as fair as possible, we assign the same structure and hidden size for all classifiers. GAIN has exactly the same structure in the generator and the same number of hidden layers in the discriminator as Classifier-GAIN.  All of the networks are designed as multi-layer perceptrons with two hidden layers. 
We use batch normalization to normalize the input layer by re-centering and re-scaling. ReLU activation function and dropout are applied after each hidden linear layer. All of the neural networks utilize Sigmoid activation at the last step for outputs. The hidden layer settings in all of our experiments are listed in Table~\ref{tab:config }.

We implement our model and its variants using PyTorch~\cite{paszke2019pytorch}, and use a GeForce GTX TITAN X 12 GB GPU for training, validation as well as testing. 
All of the neural networks are trained by using the Adam optimizer \cite{kingma2014adam}, whose learning rates are selected by grid search from 0.0005 to 0.002\footnote{Other hyper-parameters are detailed in the Appendix B, and all of them are selected by grid search.}.For the convergence of the MICE imputation, we apply the IterativeImputer in Scikit-learn \cite{pedregosa2011scikit} with mean initial strategy, 100 maximum number of imputation rounds and 0.001 as tolerance of the stopping condition.
\subsection{Performance comparison}\label{sec:pc}
We conduct each experiment by running 5 times with different random initializations and show the results in the format "mean ± standard deviation" to answer Q1 and Q2. For readers' convenience, we make the best performance bold in each of the performance tables in this section.

\subsubsection*{\textbf{Synthetic data}}
To evaluate Classifier-GAIN’s capability to capture the relationship between clinical records (vital signs and laboratory values) and label outcome for downstream prediction, we randomly remove $20\%$, $25\%$, $30\%$, $35\%$, $40\%$, $45\%$ and $50\%$ of all components from  clinical records, to simulate missingness resulting from the  urgency of the clinical situation. We demonstrate the effectiveness of Classifier-GAIN against other baselines in Table~\ref{tab:sepsisprof}. To 
understand the performance gap between different missing scenarios and the completed data, we train a binary classifier on the completed dataset, which we refer to as the upper bound.  As shown in Table~\ref{tab:sepsisprof}, Classifier-GAIN consistently outperforms the simple imputation, MICE and GAIN across the entire range of missing rates, for both evaluation metrics.
Especially, when the missing rate is  $25\%$, Classifier-GAIN improves $5.2\%$ and $14.6\%$ in AUC-ROC and macro F1-score, respectively, compared with the best baselines. 

To quantitatively evaluate the performance of Classifier-GAIN, we derive two additional metrics: (1) the relative improvement rate (RIR),
\begin{equation}
\frac{\text{Classifier-GAIN  \textbf{-} best\_baseline} }{\text{best\_baseline}},
\label{eq:improvement}
\end{equation}
to demonstrate how much Classifier-GAIN improves compared to the best baseline, and (2) the relative gap reduction rate (RGRR),
\begin{equation}
\frac{\text{Classifier-GAIN  \textbf{-} best\_baseline} }{\text{Upper bound  \textbf{-} best\_baseline}},
\end{equation}
to measure the capability of Classifier-GAIN reducing the performance gap to the upper bound.

\vspace*{-\baselineskip}
\begin{table*}[h!]
\renewcommand\thetable{7} 
\centering
\setlength{\tabcolsep}{.4pt}
\begin{tabular}{ c c| c |c |c || c| c| c| c  } 
\hline
     \multirow{3}{*}{\makecell{Missing \\ Rate}}
     &  \multicolumn{4}{c|| }{ macro F1-score }  &  \multicolumn{4}{c}{ AUC-ROC}\\
   \cline{2-9}
  {} & Classifier-GAIN  &  \makecell{Simple \\ imputation} & MICE& GAIN & Classifier-GAIN  &  \makecell{Simple \\ imputation} & MICE& GAIN\\
  \hline
  $20\%$  & \textbf{68.3} $\pm$ \textbf{0.4} &64.9 $\pm$ 1.6 &65.2 $\pm$ 1.5 & 66.0 $\pm$ 0.9  & 88.9 $\pm$ 0.5& 88.2 $\pm$ 0.6 &\textbf{89.1} $\pm$ \textbf{0.2} & 88.7 $\pm$ 0.5\\
  $25\%$  &\textbf{67.9} $\pm$ \textbf{1.0} &64.3 $\pm$ 0.5& 61.5 $\pm$ 1.7 &64.6 $\pm$ 1.7 & \textbf{88.5} $\pm$ \textbf{0.4}  & 87.4 $\pm$ 0.5 & 88.4 $\pm$ 0.2 &87.8 $\pm$ 0.3\\
  $30\%$ &\textbf{70.2} $\pm$ \textbf{0.7} & 68.2 $\pm$ 1.3 & 61.6 $\pm$ 2.2 & 65.9 $\pm$ 1.0 &\textbf{89.5} $\pm$ \textbf{0.5} & 89.0 $\pm$ 0.2 & 87.7 $\pm$ 0.2 &88.7 $\pm$ 0.4  \\
  $35\%$&\textbf{67.7} $\pm$ \textbf{0.8 }&64.5 $\pm$ 1.2 &60.4 $\pm$ 2.1 & 65.0 $\pm$ 0.4  &\textbf{88.5} $\pm$ \textbf{0.4}&87.7 $\pm$ 0.4 & 87.0 $\pm$ 0.6 & 87.4 $\pm$ 0.5 \\
  $40\%$ &\textbf{65.7} $\pm$ \textbf{1.4} &61.9 $\pm$ 0.8&  58.1 $\pm$ 3.0 & 63.2 $\pm$ 1.2  &\textbf{87.8} $\pm$ \textbf{0.5} &86.8 $\pm$ 0.4 & 85.3 $\pm$ 0.2 & 86.5 $\pm$ 0.5\\
  $45\%$&\textbf{65.7} $\pm$ \textbf{1.0}  &62.5 $\pm$ 1.4& 57.3 $\pm$ 2.7 & 61.7 $\pm$ 1.7 & \textbf{86.8} $\pm$ \textbf{0.6} &85.7 $\pm$ 0.7& 84.6 $\pm$ 0.2 &85.0 $\pm$ 0.7 \\
  $50\%$& \textbf{65.0} $\pm$ \textbf{0.2}& 59.4 $\pm$ 1.6 & 57.3 $\pm$ 3.3 & 60.0 $\pm$ 1.7 &  83.9 $\pm$ 0.8 & 83.7 $\pm$ 0.5&  \textbf{84.2} $\pm$ \textbf{0.1}  &83.5 $\pm$ 1.4 \\
\hline
     \bottomrule
\end{tabular}

\caption{ Model performance(\%) on UCSF MOF dataset with different additionally simulated missing ratios.}
\label{tab:mofprof}
\end{table*}

\begin{figure}[h]
    \centering
    \subfloat[The relative improvement rate ($\%$) ]{\includegraphics[width=0.23\textwidth,]{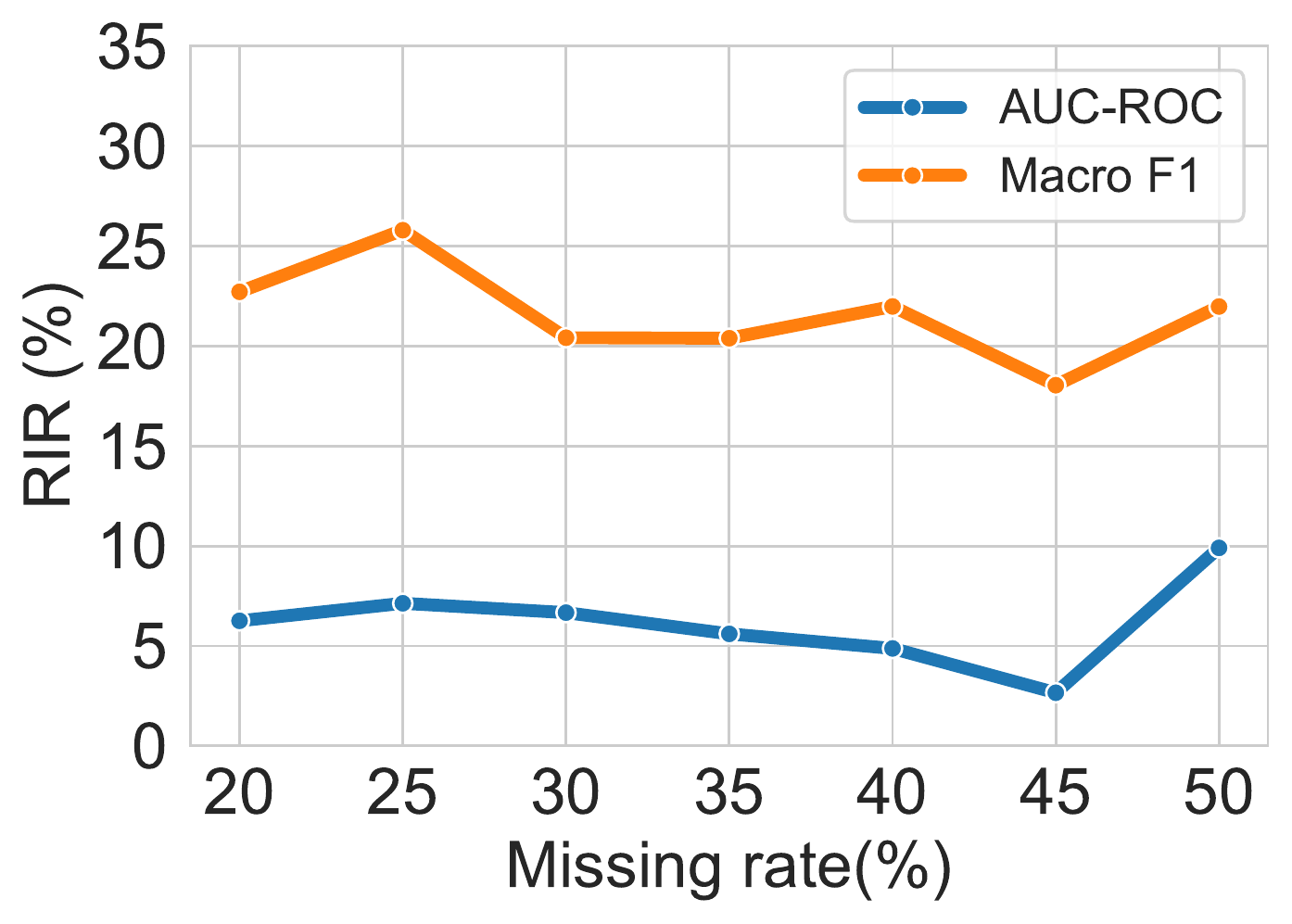}} 
    \hfill
    \subfloat[The relative gap reduction rate ($\%$)]{\includegraphics[width=0.23\textwidth, ]{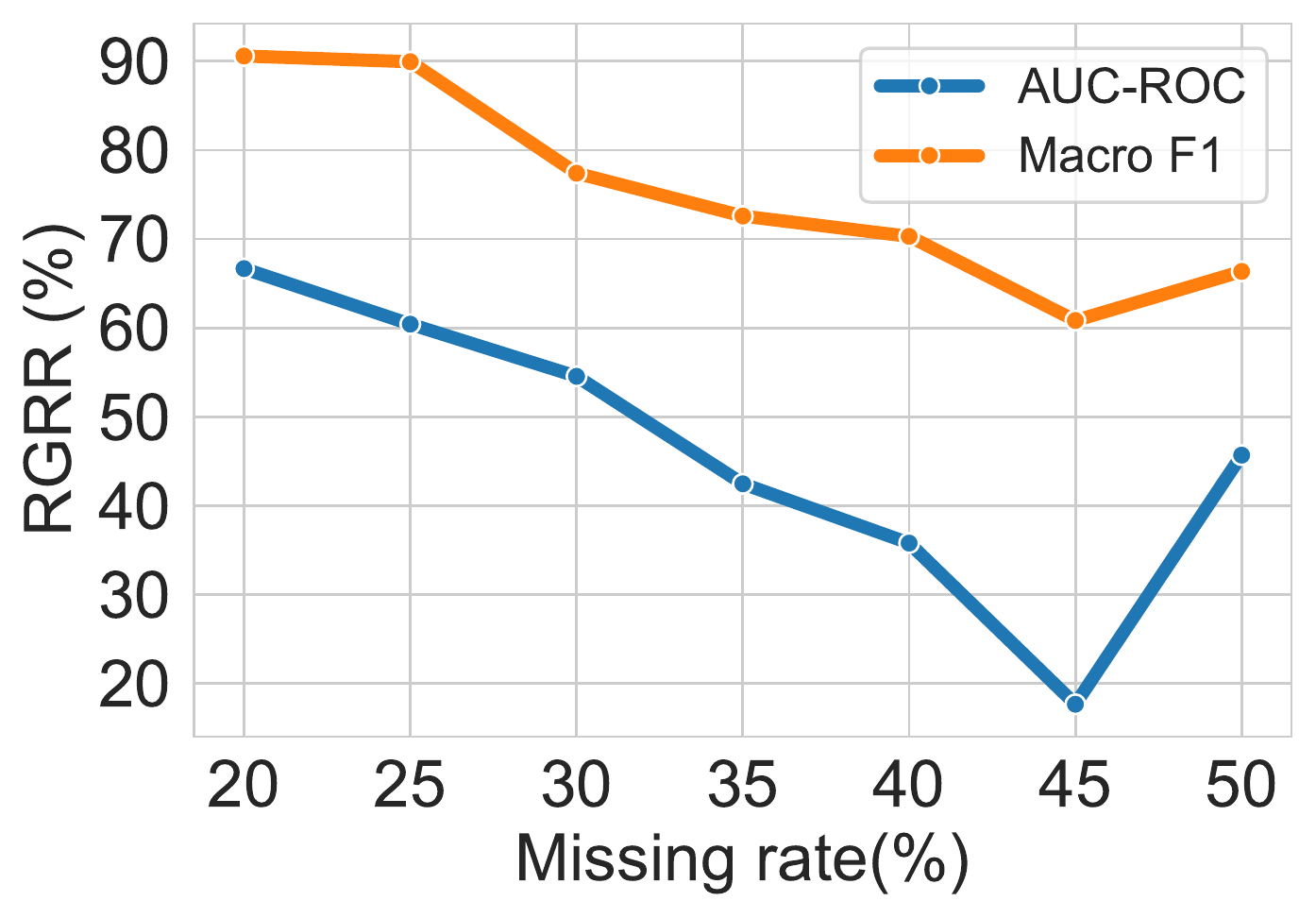}}
  \caption{The relative improvement rate (left) and the relative gap reduction rate (right) of Classifier-GAIN on AUC-ROC and macro F1-score for PhysioNet sepsis dataset across different missing ratio scenarios.
 }

  \label{fig:Simprove}
\end{figure}

The relative improvement rates calculated by Eq.~\ref{eq:improvement} across different settings are shown in Figure ~\ref{fig:Simprove} (a). For both macro F1-score and AUC-ROC, Classifier-GAIN consistently achieves a high relative improvement rate, with $21.62\%$ and $6.16\%$ on average across different scenarios, respectively. 
Especially, the relative improvement rate of macro F1-score is $25.80\%$ when the missing rate is $25\%$, and the relative improvement rate of AUC-ROC is $9.91\%$ when the missing rate is $50\%$.
Figure ~\ref{fig:Simprove} (b) shows the relative gap reduction rate of Classifier-GAIN with different missing ratio settings. Classifier-GAIN significantly reduces 
the performance gap to the upper bound, with a 75.43\% relative reduction rate for macro F1-score on average compared to best baselines, making the prediction less susceptible to missingness across different scenarios. Especially when missing rates are 20\% and 25\% (relatively low), the relative gap reduction rates are as high as 90.58\% and 89.94\%, which significantly narrows the performance gap caused by missing components. 
Even when missing rates are 40\% and 50\% (very high), the relative gap reduction rates remain 60.82\% and 66.35\%, which further validates Classifier-GAIN's applicability in different missing scenarios. 

\subsubsection*{\textbf{Real-world data}}We further evaluate our model on the UCSF MOF real-world dataset, including early-stage clinical records for MOF prediction. In addition to the high missing ratio in bio-marker measurements, there is a serious label imbalance issue in this dataset, which is common in real-world clinical data. We evaluate the performance of Classifier-GAIN on the UCSF MOF dataset in the following settings: (1)  imputing the original missing components and predicting MOF outcome; (2) adding additional random masks with  $20\%$, $25\%$,  $30\%$,  $35\%$, $40\%$,  $45\%$ and  $50\%$ missing rates to simulate more serious missing situations in real-world data. 

Table~\ref{tab:perf} reports the macro F1-score and AUC-ROC to evaluate Classifier-GAIN's prediction performance against other methods, on the UCSF MOF dataset with original missing components (The missing ratio of features is $6.42\%$ among all patients on average.). Classifier-GAIN yields the best prediction performance as measured by both  macro F1-score and AUC-ROC. All the three baselines in the original missing scenario have similar performance, and the simple imputation has the best performance in baselines, which may due to the small size and high imbalance of the real-world data. Classifier-GAIN shows better performance reflecting the classifier-guided imputation helps the downstream MOF prediction by making imputed values label-aware.

\begin{table}[H]
\renewcommand\thetable{6} 
\centering
\setlength{\tabcolsep}{.5pt}
 \begin{tabular}{c  c  c }
 \hline
    Algorithm &  macro F1-score & AUC-ROC \\
\hline
Classifier-GAIN & \textbf{71.0} $\pm$ \textbf{1.0} &\textbf{90.6} $\pm$ \textbf{0.5}\\
 \hline
 Simple imputation  &68.9 $\pm$ 1.0  & 90.3 $\pm$ 0.3\\
 MICE  & 68.2 $\pm$  0.8 &  90.0 $\pm$ 0.4 \\
 GAIN & 68.2 $\pm$  0.9   & 90.2 $\pm$ 0.3\\
     \bottomrule
\end{tabular}

\caption{Model performance on UCSF MOF dataset with original missing components.}
\label{tab:perf}
\end{table} 
\vspace*{-0.5cm}

For more missing ratios in our simulated setting,  the corresponding macro F1-score and AUC-ROC are shown in Table~\ref{tab:mofprof}.
For the macro F1-score, Classifier-GAIN consistently outperforms the best baselines more than $2\%$ across the entire range of missing rates. Especially, in the $50\%$ missing scenario, Classifier-GAIN improves $5\%$ comparing to the best baseline, GAIN.
For AUC-ROC, Classifier-GAIN outperforms other post-imputation predictions in $25\%$, $30\%$, $35\%$, $40\%$, $45\%$ missing scenarios, and achieves comparable performance with MICE in $20\%$ and $50\%$ missing conditions.

\subsection{Imputation results}\label{case}
To further visualize the imputation outcomes of the generator in Classifier-GAIN, we compare the imputation results for the original missingness in the UCSF MOF dataset of Classifier-GAIN and other baselines. We select three features: D-Dimer, Factor VII (blood test)\footnote{In the remainder of this subsection, we use Factor VII to represent Factor VII (blood test).} and respiratory rate, for the imputation study. Both D-Dimer and Factor VII have more than $40\%$ missing in the original dataset, and the missing rate of respiratory rate is $7.31 \%$. D-Dimer is an indicator of patients who may develop organ failure in the further course of acute pancreatitis\cite{radenkovic2009d}. Factor VII and respiratory rate are highly related to pulmonary failure \cite{wei1990assessment, del2011acute}.

\begin{figure}[h]
    \centering
    \subfloat[Factor VII imputation for no MOF]{\includegraphics[width=0.23\textwidth,]{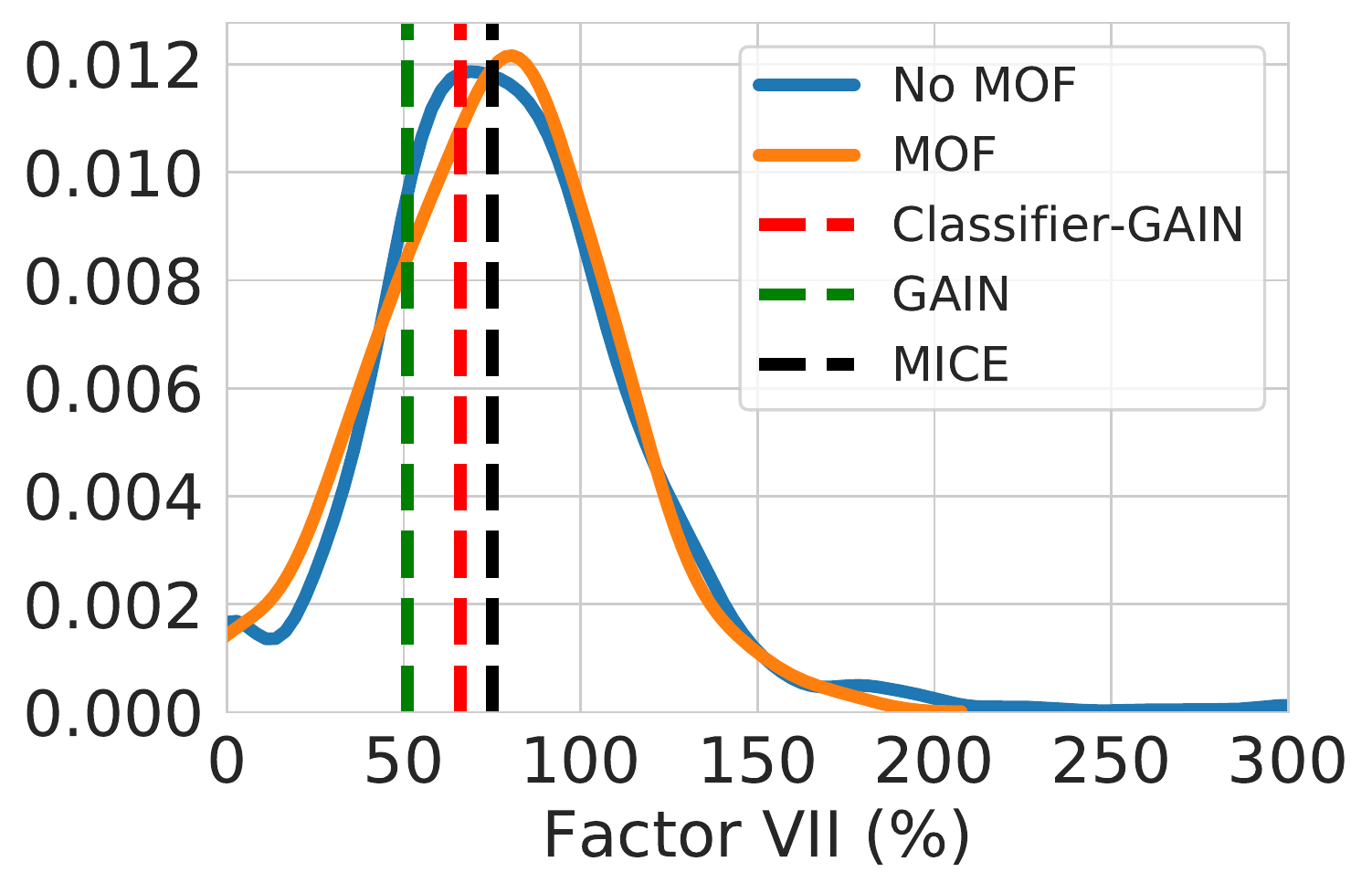}} 
    \hfill
    \subfloat[Respiratory rate imputation for no MOF]{\includegraphics[width=0.23\textwidth, ]{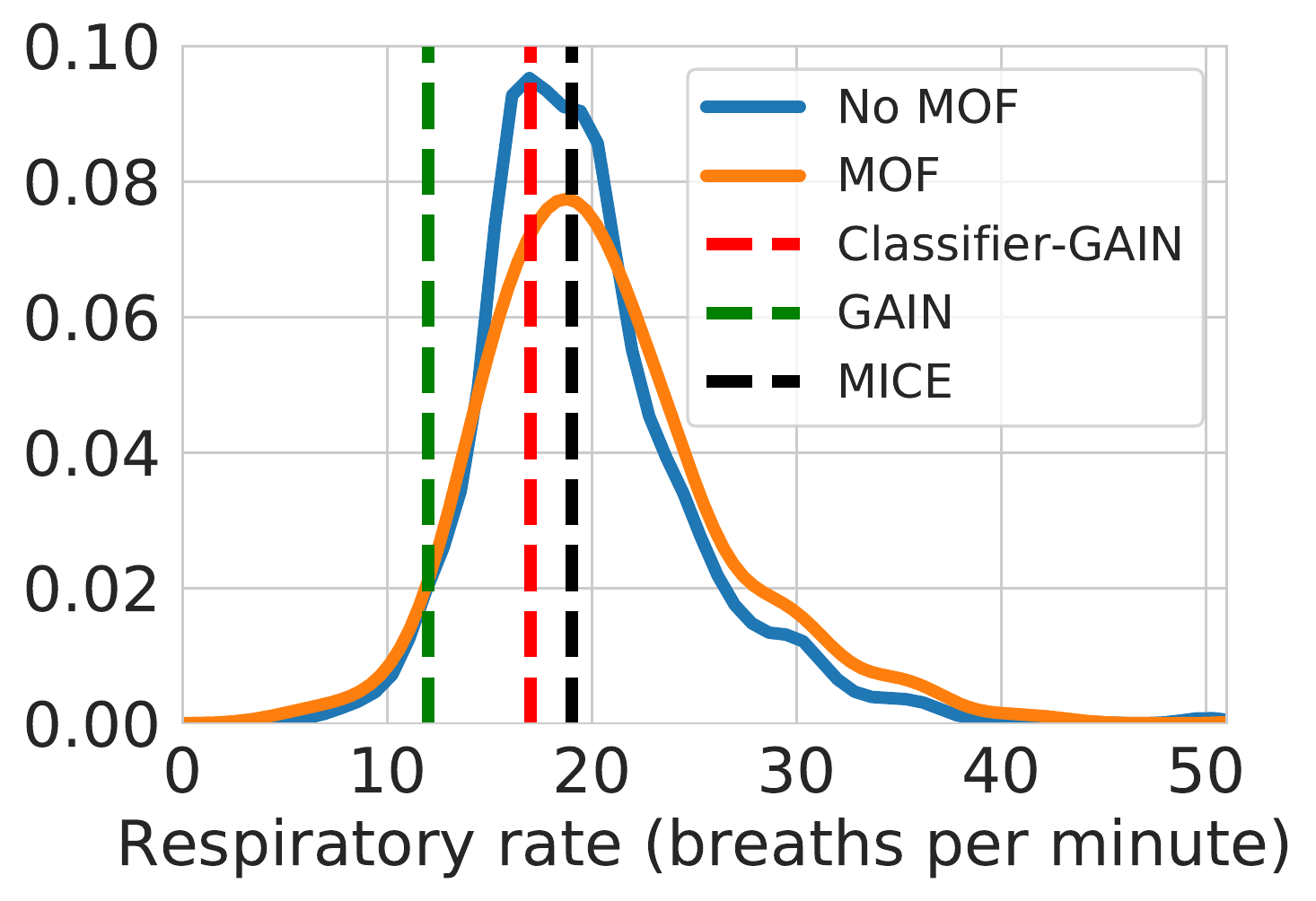}}
    \hfill
    \subfloat[D-Dimer imputation for MOF]{\includegraphics[width=0.23\textwidth, ]{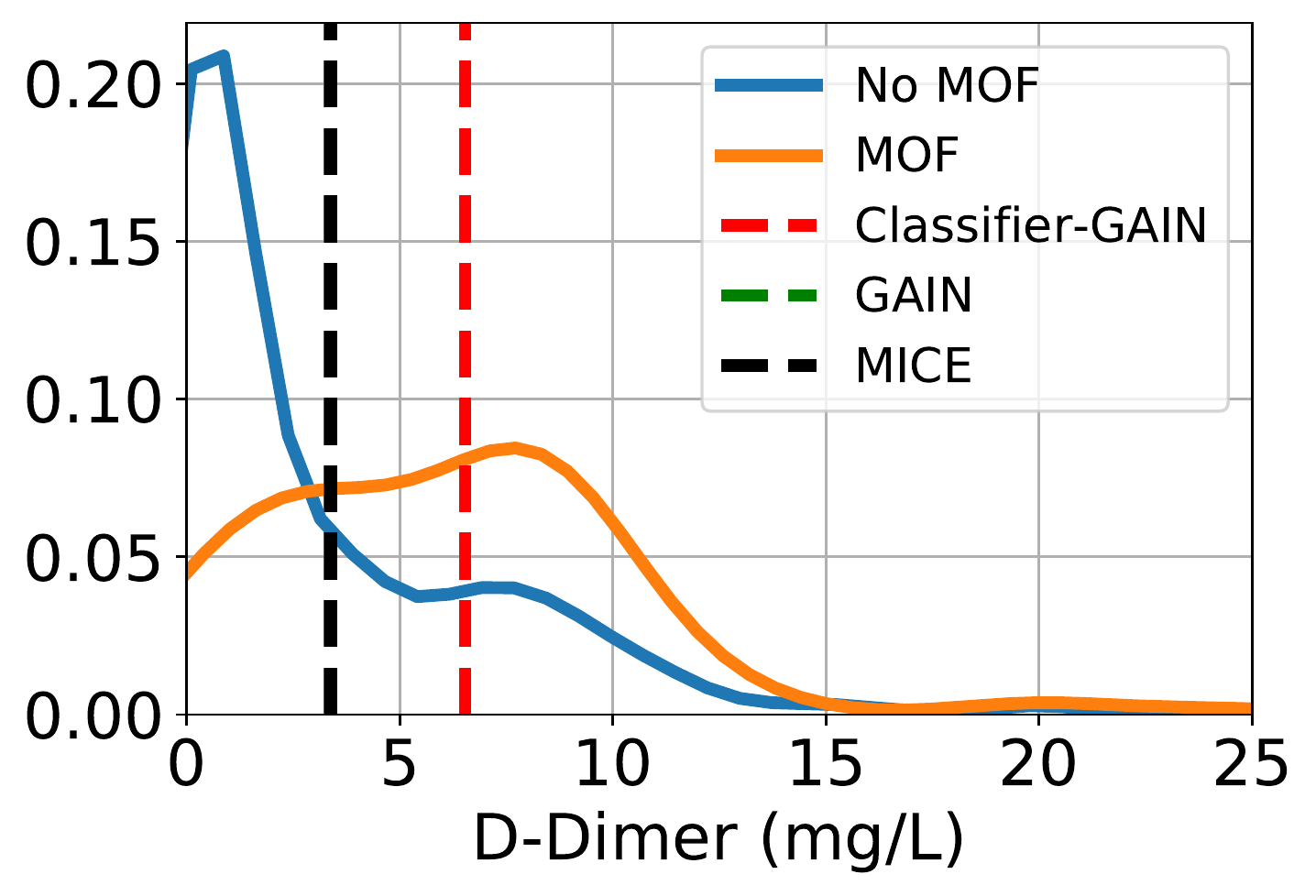}}
    \hfill
    \subfloat[Factor VII imputation for no MOF]{\includegraphics[width=0.23\textwidth, ]{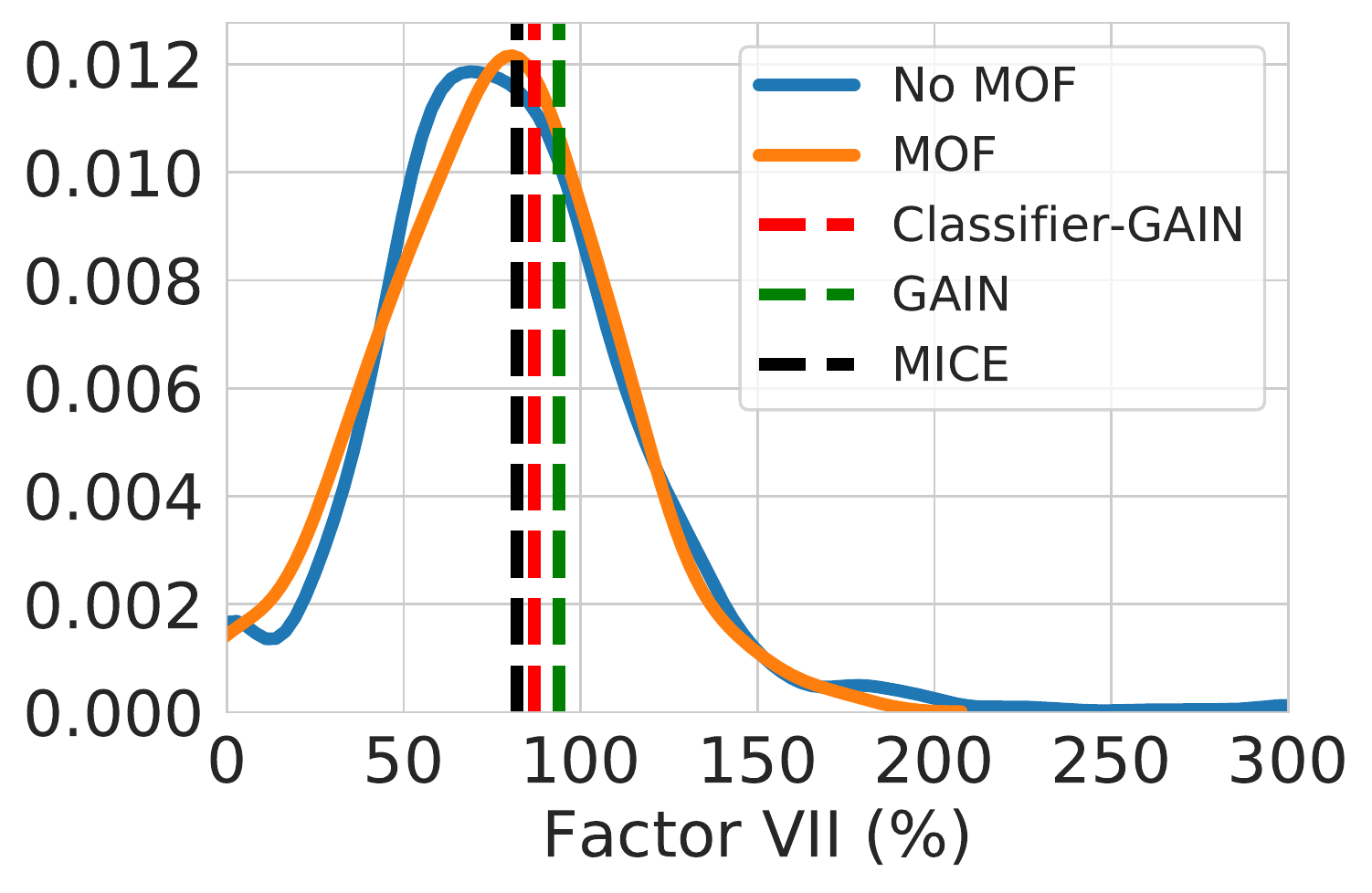}}
  \caption{Density plots of features and imputed values.
 The blue and orange curves are density curves of observed data points. The blue curve represents MOF = 0, and the orange represents MOF = 1. The dashed vertical lines are the imputation results of three different imputation methods.For (c), GAIN and MICE have similar imputation results, which makes lines overlap.}
  \Description{case.}
  \label{fig:case}
\end{figure}

Figure~\ref{fig:case} plots the univariate distributions of selected features for MOF (MOF = 1) and no MOF (MOF = 0) patients, respectively. The blue and orange curves are density curves of observed components of features that need imputation. The blue curves represent the density curves of MOF = 0, and the orange ones represent MOF = 1. Dashed vertical lines are the imputed results of three different imputation methods: red is Classifier-GAIN, green is GAIN, and black is MICE imputation. In the UCSF MOF dataset, the feature values available with maximum density of D-Dimer, Factor VII and respiratory rate for patients who did not develop MOF are $0.86$ mg/L, $68.82\%$ and $16.92$ breaths per minute, respectively. For MOF patients,  the feature values available with the maximum density are $7.71$ mg/L, $80.69\%$ and $18.75$ breaths per minute, respectively.

For panels (a) and (b), Classifier-GAIN predicts correctly for a patient without MOF, while the other classifiers, whose input data are imputed by MICE or GAIN, predict incorrectly. The imputed values of Classifier-GAIN for Factor VII and respiratory rate  are relatively closer to the feature values with maximum density for no-MOF's in both cases. Panel (c) shows a MOF patient whose data was missing the D-Dimer record.  MICE and GAIN have very similar imputed value which are $3.38$ mg/L and $3.35$ mg/L, Classifier-GAIN imputes the D-Dimer as $6.53$,mg/L which follows the trend of MOF patients in this dataset. Panel (d) shows a situation that all of the classifiers predict a no-MOF case incorrectly. In this case, all three methods impute the Factor VII closer to the feature value with the maximum density for MOF’s.
\section{related work}\label{related}

\subsubsection*{\textbf{Generative Adversarial Networks (GAN)}} 
GAN, introduced in~\cite{goodfellow2014generative}, is a game-theoretic framework for estimating the implicit distribution of data via an adversarial process. CGAN conditions the GAN framework on class labels to direct the data generation process~\cite{mirza2014conditional}. AC-GAN further improves generation performance by modifying the discriminator to contain an auxiliary decoder network~\cite{odena2017conditional}. However, both CGAN and AC-GAN need to feed label information to their generators, which could not be achieved if the final goal is classification and the label information is unknown during inference. Semi-supervised GAN~\cite{odena2016semi} performs GAN in a semi-supervised context to make the discriminator output either data validation or class labels. Triple-GAN facilitates the convergence of both the generator and the discriminator by introducing the "third player" -- classifier~\cite{li2017triple}.

Researchers have also applied GANs on missing value imputation. In
GAIN~\cite{yoon2018gain}, the generator imputes the missing components while the discriminator takes a completed vector and attempts to determine which components were actually observed and which were imputed with some additional information in the form of a hint vector. MISGAN learns a complete data generator along with a mask generator that models the missing data distribution and an adversarially trained imputer~\cite{li2019misgan}. However, those existing methods  ignore the connection between observations and classification
information, which can make use for learning label-aware imputation during training and help to improve downstream task prediction during inference.

\subsubsection*{\textbf{Multiple Organ Failure (MOF)}}

MOF is a major threat to the survival of patients with sepsis and is becoming the most common cause of death for surgical ICU patients~\cite{brown2006neutrophils}. According to a recent study  of  ICU  trauma  patients,  almost half of them developed MOF, and MOF increased the overall risk of death 6.0 times compared to the patients without MOF~\cite{ulvik2007multiple}. Sepsis is viewed as an immune storm that leads to MOF and death, which still is a leading cause of death in critically ill patients, though modern antibiotics and new resuscitation therapies have been used~\cite{gustot2011multiple}. The
Acute Physiology and Chronic Health Evaluation (APACHE) score and the Ranson score are widely used for seriously ill patients, but their empirical utilization for predicting the risk of MOF at an early stage is limited by cumbersomeness and needs to record some indexes dynamically~\cite{qiu2019development}. Therefore, a prognostic tool that can reliably predict MOF in the early phase is essential for improving patient outcomes.
In this work, we have chosen to base our MOF prediction on highly-related vital signs at the initial stage, to predict outcomes with classifier-guided imputation, in order to handle the data sparsity problem.

\subsubsection*{\textbf{Missing Data Mechanisms}}

Depending on the underlying reasons,  missingness is divided into three categories: missing completely at random (MCAR), missing at random (MAR), and missing not at random (MNAR).  MCAR refers to a situation in which the occurrence for a data point to be missing is entirely random. MAR assumes that the missingness does not have any relationship with the missing data but may depend on the observed data. MNAR indicates that the missing elements are related to the reasons for which the data is missing. In general, we assume that the EHR data is MAR data because, in most EHR instances, those collected features would be expected to explain some, but not all, of the variation among patients whose data have missing values\cite{wells2013strategies}. 

Various methodologies are available to address the missing data problem.
Single imputation algorithms only impute missing components in one iteration, which can utilize some unique numbers (e.g., 0) or statistical characteristics, such as mean value imputation~\cite{acuna2004treatment}, median imputation \cite{kantardzic2011data} and most common value imputation \cite{harrell2015regression}. MICE~\cite{MICE} is one of the most commonly used multiple imputation algorithms, applying multiple regression models iteratively to impute missing values for different types of variables \cite{zhang2020missing}. Grape is a graph-based framework with both feature imputation and label prediction, which formulates missing components imputation as an edge-level prediction and downstream label prediction as a node-level prediction \cite{you2020handling}. Unlike our work, Grape predicts the downstream task without caring about imputed data and the feature imputation only learns information from partially observed data, which is not label-aware. In this work, we have explored the algorithm with missing components in EHRs datasets to resolve the real-world MOF prediction task.

\section{Conclusion}\label{discussion}
In this paper, we present Classifier-GAIN, an end-to-end deep learning framework to improve performance of MOF prediction on datasets with a wide range of missingness ratios. In contrast to most of the label-aware GANs, whose generator takes label information directly, focusing on improving the generator outputs, we design a three-player adversarial imputation network to optimize the downstream prediction while imputing missing values. Classifier-GAIN uses a classifier to provide label supervision signals to the generator in training, and the trained generator to improve the classifier’s downstream prediction performance in inference.
Extensive experimental results on both a synthetic sepsis dataset and a real world MOF dataset demonstrate the usefulness of this framework. Although we only demonstrate the effeteness of Classifier-GAIN in MOF prediction tasks, its applications in other domains are worth exploring, which we leave as further work.


\begin{acks}
This work was funded by the National Institutes for Health (NIH) grant
NIH 7R01HL149670.
\end{acks}

\bibliographystyle{ACM-Reference-Format}
\bibliography{ref}


\begin{thebibliography}{34}


\ifx \showCODEN    \undefined \def \showCODEN     #1{\unskip}     \fi
\ifx \showDOI      \undefined \def \showDOI       #1{#1}\fi
\ifx \showISBNx    \undefined \def \showISBNx     #1{\unskip}     \fi
\ifx \showISBNxiii \undefined \def \showISBNxiii  #1{\unskip}     \fi
\ifx \showISSN     \undefined \def \showISSN      #1{\unskip}     \fi
\ifx \showLCCN     \undefined \def \showLCCN      #1{\unskip}     \fi
\ifx \shownote     \undefined \def \shownote      #1{#1}          \fi
\ifx \showarticletitle \undefined \def \showarticletitle #1{#1}   \fi
\ifx \showURL      \undefined \def \showURL       {\relax}        \fi
\providecommand\bibfield[2]{#2}
\providecommand\bibinfo[2]{#2}
\providecommand\natexlab[1]{#1}
\providecommand\showeprint[2][]{arXiv:#2}

\bibitem[\protect\citeauthoryear{Acuna and Rodriguez}{Acuna and
  Rodriguez}{2004}]%
        {acuna2004treatment}
\bibfield{author}{\bibinfo{person}{Edgar Acuna} {and} \bibinfo{person}{Caroline
  Rodriguez}.} \bibinfo{year}{2004}\natexlab{}.
\newblock \showarticletitle{The treatment of missing values and its effect on
  classifier accuracy}.
\newblock In \bibinfo{booktitle}{\emph{Classification, clustering, and data
  mining applications}}. \bibinfo{publisher}{Springer},
  \bibinfo{pages}{639--647}.
\newblock


\bibitem[\protect\citeauthoryear{Bakker, Gris, Coffernils, Kahn, and
  Vincent}{Bakker et~al\mbox{.}}{1996}]%
        {bakker1996serial}
\bibfield{author}{\bibinfo{person}{Jan Bakker}, \bibinfo{person}{Philippe
  Gris}, \bibinfo{person}{Michel Coffernils}, \bibinfo{person}{Robert~J Kahn},
  {and} \bibinfo{person}{Jean-Louis Vincent}.} \bibinfo{year}{1996}\natexlab{}.
\newblock \showarticletitle{Serial blood lactate levels can predict the
  development of multiple organ failure following septic shock}.
\newblock \bibinfo{journal}{\emph{The American journal of surgery}}
  \bibinfo{volume}{171}, \bibinfo{number}{2} (\bibinfo{year}{1996}),
  \bibinfo{pages}{221--226}.
\newblock


\bibitem[\protect\citeauthoryear{Bravo-Merodio, Acharjee, Hazeldine, Bentley,
  Foster, Gkoutos, and Lord}{Bravo-Merodio et~al\mbox{.}}{2019}]%
        {bravo2019machine}
\bibfield{author}{\bibinfo{person}{Laura Bravo-Merodio},
  \bibinfo{person}{Animesh Acharjee}, \bibinfo{person}{Jon Hazeldine},
  \bibinfo{person}{Conor Bentley}, \bibinfo{person}{Mark Foster},
  \bibinfo{person}{Georgios~V Gkoutos}, {and} \bibinfo{person}{Janet~M Lord}.}
  \bibinfo{year}{2019}\natexlab{}.
\newblock \showarticletitle{Machine learning for the detection of early
  immunological markers as predictors of multi-organ dysfunction}.
\newblock \bibinfo{journal}{\emph{Scientific data}} \bibinfo{volume}{6},
  \bibinfo{number}{1} (\bibinfo{year}{2019}), \bibinfo{pages}{1--10}.
\newblock


\bibitem[\protect\citeauthoryear{Brown, Brain, Pearson, Edgeworth, Lewis, and
  Treacher}{Brown et~al\mbox{.}}{2006}]%
        {brown2006neutrophils}
\bibfield{author}{\bibinfo{person}{KA Brown}, \bibinfo{person}{SD Brain},
  \bibinfo{person}{JD Pearson}, \bibinfo{person}{JD Edgeworth},
  \bibinfo{person}{SM Lewis}, {and} \bibinfo{person}{DF Treacher}.}
  \bibinfo{year}{2006}\natexlab{}.
\newblock \showarticletitle{Neutrophils in development of multiple organ
  failure in sepsis}.
\newblock \bibinfo{journal}{\emph{The Lancet}} \bibinfo{volume}{368},
  \bibinfo{number}{9530} (\bibinfo{year}{2006}), \bibinfo{pages}{157--169}.
\newblock


\bibitem[\protect\citeauthoryear{Buuren and Groothuis-Oudshoorn}{Buuren and
  Groothuis-Oudshoorn}{2010}]%
        {MICE}
\bibfield{author}{\bibinfo{person}{S~van Buuren} {and} \bibinfo{person}{Karin
  Groothuis-Oudshoorn}.} \bibinfo{year}{2010}\natexlab{}.
\newblock \showarticletitle{mice: Multivariate imputation by chained equations
  in R}.
\newblock \bibinfo{journal}{\emph{Journal of statistical software}}
  (\bibinfo{year}{2010}), \bibinfo{pages}{1--68}.
\newblock


\bibitem[\protect\citeauthoryear{Del~Sorbo and Slutsky}{Del~Sorbo and
  Slutsky}{2011}]%
        {del2011acute}
\bibfield{author}{\bibinfo{person}{Lorenzo Del~Sorbo} {and}
  \bibinfo{person}{Arthur~S Slutsky}.} \bibinfo{year}{2011}\natexlab{}.
\newblock \showarticletitle{Acute Respiratory Distress Syndrome and Multiple
  Organ Failure}.
\newblock \bibinfo{journal}{\emph{Current opinion in critical care}}
  \bibinfo{volume}{17}, \bibinfo{number}{1} (\bibinfo{year}{2011}),
  \bibinfo{pages}{1--6}.
\newblock


\bibitem[\protect\citeauthoryear{Durham, Moran, Mazuski, Shapiro, Baue, and
  Flint}{Durham et~al\mbox{.}}{2003}]%
        {durham2003multiple}
\bibfield{author}{\bibinfo{person}{Rodney~M Durham}, \bibinfo{person}{JJ
  Moran}, \bibinfo{person}{John~E Mazuski}, \bibinfo{person}{Marc~J Shapiro},
  \bibinfo{person}{Arthur~E Baue}, {and} \bibinfo{person}{Lewis~M Flint}.}
  \bibinfo{year}{2003}\natexlab{}.
\newblock \showarticletitle{Multiple organ failure in trauma patients}.
\newblock \bibinfo{journal}{\emph{Journal of Trauma and Acute Care Surgery}}
  \bibinfo{volume}{55}, \bibinfo{number}{4} (\bibinfo{year}{2003}),
  \bibinfo{pages}{608--616}.
\newblock


\bibitem[\protect\citeauthoryear{Goodfellow, Pouget-Abadie, Mirza, Xu,
  Warde-Farley, Ozair, Courville, and Bengio}{Goodfellow et~al\mbox{.}}{2014}]%
        {goodfellow2014generative}
\bibfield{author}{\bibinfo{person}{Ian Goodfellow}, \bibinfo{person}{Jean
  Pouget-Abadie}, \bibinfo{person}{Mehdi Mirza}, \bibinfo{person}{Bing Xu},
  \bibinfo{person}{David Warde-Farley}, \bibinfo{person}{Sherjil Ozair},
  \bibinfo{person}{Aaron Courville}, {and} \bibinfo{person}{Yoshua Bengio}.}
  \bibinfo{year}{2014}\natexlab{}.
\newblock \showarticletitle{Generative adversarial nets}.
\newblock \bibinfo{journal}{\emph{Advances in neural information processing
  systems}}  \bibinfo{volume}{27} (\bibinfo{year}{2014}),
  \bibinfo{pages}{2672--2680}.
\newblock


\bibitem[\protect\citeauthoryear{Gustot}{Gustot}{2011}]%
        {gustot2011multiple}
\bibfield{author}{\bibinfo{person}{Thierry Gustot}.}
  \bibinfo{year}{2011}\natexlab{}.
\newblock \showarticletitle{Multiple organ failure in sepsis: prognosis and
  role of systemic inflammatory response}.
\newblock \bibinfo{journal}{\emph{Current opinion in critical care}}
  \bibinfo{volume}{17}, \bibinfo{number}{2} (\bibinfo{year}{2011}),
  \bibinfo{pages}{153--159}.
\newblock


\bibitem[\protect\citeauthoryear{Harjola, Mullens, Banaszewski, Bauersachs,
  Brunner-La~Rocca, Chioncel, Collins, Doehner, Filippatos, Flammer,
  et~al\mbox{.}}{Harjola et~al\mbox{.}}{2017}]%
        {harjola2017organ}
\bibfield{author}{\bibinfo{person}{Veli-Pekka Harjola},
  \bibinfo{person}{Wilfried Mullens}, \bibinfo{person}{Marek Banaszewski},
  \bibinfo{person}{Johann Bauersachs}, \bibinfo{person}{Hans-Peter
  Brunner-La~Rocca}, \bibinfo{person}{Ovidiu Chioncel}, \bibinfo{person}{Sean~P
  Collins}, \bibinfo{person}{Wolfram Doehner}, \bibinfo{person}{Gerasimos~S
  Filippatos}, \bibinfo{person}{Andreas~J Flammer}, {et~al\mbox{.}}}
  \bibinfo{year}{2017}\natexlab{}.
\newblock \showarticletitle{Organ dysfunction, injury and failure in acute
  heart failure: from pathophysiology to diagnosis and management. A review on
  behalf of the Acute Heart Failure Committee of the Heart Failure Association
  (HFA) of the European Society of Cardiology (ESC)}.
\newblock \bibinfo{journal}{\emph{European journal of heart failure}}
  \bibinfo{volume}{19}, \bibinfo{number}{7} (\bibinfo{year}{2017}),
  \bibinfo{pages}{821--836}.
\newblock


\bibitem[\protect\citeauthoryear{Harrell~Jr}{Harrell~Jr}{2015}]%
        {harrell2015regression}
\bibfield{author}{\bibinfo{person}{Frank~E Harrell~Jr}.}
  \bibinfo{year}{2015}\natexlab{}.
\newblock \bibinfo{booktitle}{\emph{Regression modeling strategies: with
  applications to linear models, logistic and ordinal regression, and survival
  analysis}}.
\newblock \bibinfo{publisher}{Springer}.
\newblock


\bibitem[\protect\citeauthoryear{Kantardzic}{Kantardzic}{2011}]%
        {kantardzic2011data}
\bibfield{author}{\bibinfo{person}{Mehmed Kantardzic}.}
  \bibinfo{year}{2011}\natexlab{}.
\newblock \bibinfo{booktitle}{\emph{Data mining: concepts, models, methods, and
  algorithms}}.
\newblock \bibinfo{publisher}{John Wiley \& Sons}.
\newblock


\bibitem[\protect\citeauthoryear{Kingma and Ba}{Kingma and Ba}{2014}]%
        {kingma2014adam}
\bibfield{author}{\bibinfo{person}{Diederik~P Kingma} {and}
  \bibinfo{person}{Jimmy Ba}.} \bibinfo{year}{2014}\natexlab{}.
\newblock \showarticletitle{Adam: A method for stochastic optimization}.
\newblock \bibinfo{journal}{\emph{arXiv preprint arXiv:1412.6980}}
  (\bibinfo{year}{2014}).
\newblock


\bibitem[\protect\citeauthoryear{Li, Xu, Zhu, and Zhang}{Li
  et~al\mbox{.}}{2017}]%
        {li2017triple}
\bibfield{author}{\bibinfo{person}{Chongxuan Li}, \bibinfo{person}{Taufik Xu},
  \bibinfo{person}{Jun Zhu}, {and} \bibinfo{person}{Bo Zhang}.}
  \bibinfo{year}{2017}\natexlab{}.
\newblock \showarticletitle{Triple generative adversarial nets}.
\newblock \bibinfo{journal}{\emph{Advances in neural information processing
  systems}}  \bibinfo{volume}{30} (\bibinfo{year}{2017}),
  \bibinfo{pages}{4088--4098}.
\newblock


\bibitem[\protect\citeauthoryear{Li, Jiang, and Marlin}{Li
  et~al\mbox{.}}{2019}]%
        {li2019misgan}
\bibfield{author}{\bibinfo{person}{Steven Cheng-Xian Li}, \bibinfo{person}{Bo
  Jiang}, {and} \bibinfo{person}{Benjamin Marlin}.}
  \bibinfo{year}{2019}\natexlab{}.
\newblock \showarticletitle{Misgan: Learning from incomplete data with
  generative adversarial networks}.
\newblock \bibinfo{journal}{\emph{arXiv preprint arXiv:1902.09599}}
  (\bibinfo{year}{2019}).
\newblock


\bibitem[\protect\citeauthoryear{Mirza and Osindero}{Mirza and
  Osindero}{2014}]%
        {mirza2014conditional}
\bibfield{author}{\bibinfo{person}{Mehdi Mirza} {and} \bibinfo{person}{Simon
  Osindero}.} \bibinfo{year}{2014}\natexlab{}.
\newblock \showarticletitle{Conditional generative adversarial nets}.
\newblock \bibinfo{journal}{\emph{arXiv preprint arXiv:1411.1784}}
  (\bibinfo{year}{2014}).
\newblock


\bibitem[\protect\citeauthoryear{Odena}{Odena}{2016}]%
        {odena2016semi}
\bibfield{author}{\bibinfo{person}{Augustus Odena}.}
  \bibinfo{year}{2016}\natexlab{}.
\newblock \showarticletitle{Semi-supervised learning with generative
  adversarial networks}.
\newblock \bibinfo{journal}{\emph{arXiv preprint arXiv:1606.01583}}
  (\bibinfo{year}{2016}).
\newblock


\bibitem[\protect\citeauthoryear{Odena, Olah, and Shlens}{Odena
  et~al\mbox{.}}{2017}]%
        {odena2017conditional}
\bibfield{author}{\bibinfo{person}{Augustus Odena},
  \bibinfo{person}{Christopher Olah}, {and} \bibinfo{person}{Jonathon Shlens}.}
  \bibinfo{year}{2017}\natexlab{}.
\newblock \showarticletitle{Conditional image synthesis with auxiliary
  classifier gans}. In \bibinfo{booktitle}{\emph{International conference on
  machine learning}}. PMLR, \bibinfo{pages}{2642--2651}.
\newblock


\bibitem[\protect\citeauthoryear{Otero-L{\'o}pez, Alonso-Hern{\'a}ndez,
  Maderuelo-Fern{\'a}ndez, Garrido-Corro, Dom{\'\i}nguez-Gil, and
  S{\'a}nchez-Rodr{\'\i}guez}{Otero-L{\'o}pez et~al\mbox{.}}{2006}]%
        {ICU}
\bibfield{author}{\bibinfo{person}{Mar{\'\i}a~Jos{\'e} Otero-L{\'o}pez},
  \bibinfo{person}{Pablo Alonso-Hern{\'a}ndez}, \bibinfo{person}{Jos{\'e}~Angel
  Maderuelo-Fern{\'a}ndez}, \bibinfo{person}{Beatriz Garrido-Corro},
  \bibinfo{person}{Alfonso Dom{\'\i}nguez-Gil}, {and} \bibinfo{person}{Angel
  S{\'a}nchez-Rodr{\'\i}guez}.} \bibinfo{year}{2006}\natexlab{}.
\newblock \showarticletitle{Preventable adverse drug events in hospitalized
  patients}.
\newblock \bibinfo{journal}{\emph{Medicina clinica}} \bibinfo{volume}{126},
  \bibinfo{number}{3} (\bibinfo{year}{2006}), \bibinfo{pages}{81--87}.
\newblock


\bibitem[\protect\citeauthoryear{Papachristou, Muddana, Yadav, O'connell,
  Sanders, Slivka, and Whitcomb}{Papachristou et~al\mbox{.}}{2010}]%
        {papachristou2010comparison}
\bibfield{author}{\bibinfo{person}{Georgios~I Papachristou},
  \bibinfo{person}{Venkata Muddana}, \bibinfo{person}{Dhiraj Yadav},
  \bibinfo{person}{Michael O'connell}, \bibinfo{person}{Michael~K Sanders},
  \bibinfo{person}{Adam Slivka}, {and} \bibinfo{person}{David~C Whitcomb}.}
  \bibinfo{year}{2010}\natexlab{}.
\newblock \showarticletitle{Comparison of BISAP, Ranson's, APACHE-II, and CTSI
  scores in predicting organ failure, complications, and mortality in acute
  pancreatitis}.
\newblock \bibinfo{journal}{\emph{American Journal of Gastroenterology}}
  \bibinfo{volume}{105}, \bibinfo{number}{2} (\bibinfo{year}{2010}),
  \bibinfo{pages}{435--441}.
\newblock


\bibitem[\protect\citeauthoryear{Paszke, Gross, Massa, Lerer, Bradbury, Chanan,
  Killeen, Lin, Gimelshein, Antiga, et~al\mbox{.}}{Paszke
  et~al\mbox{.}}{2019}]%
        {paszke2019pytorch}
\bibfield{author}{\bibinfo{person}{Adam Paszke}, \bibinfo{person}{Sam Gross},
  \bibinfo{person}{Francisco Massa}, \bibinfo{person}{Adam Lerer},
  \bibinfo{person}{James Bradbury}, \bibinfo{person}{Gregory Chanan},
  \bibinfo{person}{Trevor Killeen}, \bibinfo{person}{Zeming Lin},
  \bibinfo{person}{Natalia Gimelshein}, \bibinfo{person}{Luca Antiga},
  {et~al\mbox{.}}} \bibinfo{year}{2019}\natexlab{}.
\newblock \showarticletitle{Pytorch: An imperative style, high-performance deep
  learning library}. In \bibinfo{booktitle}{\emph{Advances in neural
  information processing systems}}. \bibinfo{pages}{8026--8037}.
\newblock


\bibitem[\protect\citeauthoryear{Pedregosa, Varoquaux, Gramfort, Michel,
  Thirion, Grisel, Blondel, Prettenhofer, Weiss, Dubourg,
  et~al\mbox{.}}{Pedregosa et~al\mbox{.}}{2011}]%
        {pedregosa2011scikit}
\bibfield{author}{\bibinfo{person}{Fabian Pedregosa}, \bibinfo{person}{Ga{\"e}l
  Varoquaux}, \bibinfo{person}{Alexandre Gramfort}, \bibinfo{person}{Vincent
  Michel}, \bibinfo{person}{Bertrand Thirion}, \bibinfo{person}{Olivier
  Grisel}, \bibinfo{person}{Mathieu Blondel}, \bibinfo{person}{Peter
  Prettenhofer}, \bibinfo{person}{Ron Weiss}, \bibinfo{person}{Vincent
  Dubourg}, {et~al\mbox{.}}} \bibinfo{year}{2011}\natexlab{}.
\newblock \showarticletitle{Scikit-learn: Machine learning in Python}.
\newblock \bibinfo{journal}{\emph{the Journal of machine Learning research}}
  \bibinfo{volume}{12} (\bibinfo{year}{2011}), \bibinfo{pages}{2825--2830}.
\newblock


\bibitem[\protect\citeauthoryear{Qiu, Nian, Guo, Tang, Lu, Wen, Wang, Chen, and
  Liu}{Qiu et~al\mbox{.}}{2019}]%
        {qiu2019development}
\bibfield{author}{\bibinfo{person}{Qiu Qiu}, \bibinfo{person}{Yong-jian Nian},
  \bibinfo{person}{Yan Guo}, \bibinfo{person}{Liang Tang}, \bibinfo{person}{Nan
  Lu}, \bibinfo{person}{Liang-zhi Wen}, \bibinfo{person}{Bin Wang},
  \bibinfo{person}{Dong-feng Chen}, {and} \bibinfo{person}{Kai-jun Liu}.}
  \bibinfo{year}{2019}\natexlab{}.
\newblock \showarticletitle{Development and validation of three
  machine-learning models for predicting multiple organ failure in moderately
  severe and severe acute pancreatitis}.
\newblock \bibinfo{journal}{\emph{BMC gastroenterology}} \bibinfo{volume}{19},
  \bibinfo{number}{1} (\bibinfo{year}{2019}), \bibinfo{pages}{1--9}.
\newblock


\bibitem[\protect\citeauthoryear{Radenkovic, Bajec, Ivancevic, Milic,
  Bumbasirevic, Jeremic, Djukic, Stefanovic, Stefanovic, Milosevic-Zbutega,
  et~al\mbox{.}}{Radenkovic et~al\mbox{.}}{2009}]%
        {radenkovic2009d}
\bibfield{author}{\bibinfo{person}{Dejan Radenkovic}, \bibinfo{person}{Djordje
  Bajec}, \bibinfo{person}{Nenad Ivancevic}, \bibinfo{person}{Natasa Milic},
  \bibinfo{person}{Vesna Bumbasirevic}, \bibinfo{person}{Vasilije Jeremic},
  \bibinfo{person}{Vladimir Djukic}, \bibinfo{person}{Branislava Stefanovic},
  \bibinfo{person}{Branislav Stefanovic}, \bibinfo{person}{Gorica
  Milosevic-Zbutega}, {et~al\mbox{.}}} \bibinfo{year}{2009}\natexlab{}.
\newblock \showarticletitle{D-dimer in acute pancreatitis: a new approach for
  an early assessment of organ failure}.
\newblock \bibinfo{journal}{\emph{Pancreas}} \bibinfo{volume}{38},
  \bibinfo{number}{6} (\bibinfo{year}{2009}), \bibinfo{pages}{655--660}.
\newblock


\bibitem[\protect\citeauthoryear{Reyna, Josef, Seyedi, Jeter, Shashikumar,
  Westover, Sharma, Nemati, and Clifford}{Reyna et~al\mbox{.}}{2019}]%
        {reyna2019early}
\bibfield{author}{\bibinfo{person}{Matthew~A Reyna}, \bibinfo{person}{Chris
  Josef}, \bibinfo{person}{Salman Seyedi}, \bibinfo{person}{Russell Jeter},
  \bibinfo{person}{Supreeth~P Shashikumar}, \bibinfo{person}{M~Brandon
  Westover}, \bibinfo{person}{Ashish Sharma}, \bibinfo{person}{Shamim Nemati},
  {and} \bibinfo{person}{Gari~D Clifford}.} \bibinfo{year}{2019}\natexlab{}.
\newblock \showarticletitle{Early prediction of sepsis from clinical data: the
  PhysioNet/Computing in Cardiology Challenge 2019}. In
  \bibinfo{booktitle}{\emph{2019 Computing in Cardiology (CinC)}}. IEEE,
  \bibinfo{pages}{Page--1}.
\newblock


\bibitem[\protect\citeauthoryear{Rossaint and Zarbock}{Rossaint and
  Zarbock}{2015}]%
        {rossaint2015pathogenesis}
\bibfield{author}{\bibinfo{person}{Jan Rossaint} {and}
  \bibinfo{person}{Alexander Zarbock}.} \bibinfo{year}{2015}\natexlab{}.
\newblock \showarticletitle{Pathogenesis of multiple organ failure in sepsis}.
\newblock \bibinfo{journal}{\emph{Critical Reviews™ in Immunology}}
  \bibinfo{volume}{35}, \bibinfo{number}{4} (\bibinfo{year}{2015}).
\newblock


\bibitem[\protect\citeauthoryear{Ulvik, Kv{\aa}le, Wentzel-Larsen, and
  Flaatten}{Ulvik et~al\mbox{.}}{2007}]%
        {ulvik2007multiple}
\bibfield{author}{\bibinfo{person}{Atle Ulvik}, \bibinfo{person}{Reidar
  Kv{\aa}le}, \bibinfo{person}{Tore Wentzel-Larsen}, {and}
  \bibinfo{person}{Hans Flaatten}.} \bibinfo{year}{2007}\natexlab{}.
\newblock \showarticletitle{Multiple organ failure after trauma affects even
  long-term survival and functional status}.
\newblock \bibinfo{journal}{\emph{Critical Care}} \bibinfo{volume}{11},
  \bibinfo{number}{5} (\bibinfo{year}{2007}), \bibinfo{pages}{R95}.
\newblock


\bibitem[\protect\citeauthoryear{Wang, Chen, Wang, Li, Zhu, Huang, Chen, Chen,
  Deng, Lan, et~al\mbox{.}}{Wang et~al\mbox{.}}{2018}]%
        {wang2018clinical}
\bibfield{author}{\bibinfo{person}{Zhan-Ke Wang}, \bibinfo{person}{Rong-Jian
  Chen}, \bibinfo{person}{Shi-Liang Wang}, \bibinfo{person}{Guang-Wei Li},
  \bibinfo{person}{Zhong-Zhen Zhu}, \bibinfo{person}{Qiang Huang},
  \bibinfo{person}{Zi-Li Chen}, \bibinfo{person}{Fan-Chang Chen},
  \bibinfo{person}{Lei Deng}, \bibinfo{person}{Xiao-Peng Lan}, {et~al\mbox{.}}}
  \bibinfo{year}{2018}\natexlab{}.
\newblock \showarticletitle{Clinical application of a novel diagnostic scheme
  including pancreatic $\beta$-cell dysfunction for traumatic multiple organ
  dysfunction syndrome}.
\newblock \bibinfo{journal}{\emph{Molecular Medicine Reports}}
  \bibinfo{volume}{17}, \bibinfo{number}{1} (\bibinfo{year}{2018}),
  \bibinfo{pages}{683--693}.
\newblock


\bibitem[\protect\citeauthoryear{Wei}{Wei}{1990}]%
        {wei1990assessment}
\bibfield{author}{\bibinfo{person}{SJ Wei}.} \bibinfo{year}{1990}\natexlab{}.
\newblock \showarticletitle{The assessment of factor VIII-related antigen in
  endothelial cells of pulmonary blood vessels in multiple organ failure}.
\newblock \bibinfo{journal}{\emph{Zhonghua jie he he hu xi za zhi= Zhonghua
  jiehe he huxi zazhi= Chinese journal of tuberculosis and respiratory
  diseases}} \bibinfo{volume}{13}, \bibinfo{number}{6} (\bibinfo{year}{1990}),
  \bibinfo{pages}{346--8}.
\newblock


\bibitem[\protect\citeauthoryear{Wells, Chagin, Nowacki, and Kattan}{Wells
  et~al\mbox{.}}{2013}]%
        {wells2013strategies}
\bibfield{author}{\bibinfo{person}{Brian~J Wells}, \bibinfo{person}{Kevin~M
  Chagin}, \bibinfo{person}{Amy~S Nowacki}, {and} \bibinfo{person}{Michael~W
  Kattan}.} \bibinfo{year}{2013}\natexlab{}.
\newblock \showarticletitle{Strategies for handling missing data in electronic
  health record derived data}.
\newblock \bibinfo{journal}{\emph{Egems}} \bibinfo{volume}{1},
  \bibinfo{number}{3} (\bibinfo{year}{2013}).
\newblock


\bibitem[\protect\citeauthoryear{Yoon, Jordon, and Van Der~Schaar}{Yoon
  et~al\mbox{.}}{2018}]%
        {yoon2018gain}
\bibfield{author}{\bibinfo{person}{Jinsung Yoon}, \bibinfo{person}{James
  Jordon}, {and} \bibinfo{person}{Mihaela Van Der~Schaar}.}
  \bibinfo{year}{2018}\natexlab{}.
\newblock \showarticletitle{Gain: Missing data imputation using generative
  adversarial nets}.
\newblock \bibinfo{journal}{\emph{arXiv preprint arXiv:1806.02920}}
  (\bibinfo{year}{2018}).
\newblock


\bibitem[\protect\citeauthoryear{You, Ma, Ding, Kochenderfer, and Leskovec}{You
  et~al\mbox{.}}{2020}]%
        {you2020handling}
\bibfield{author}{\bibinfo{person}{Jiaxuan You}, \bibinfo{person}{Xiaobai Ma},
  \bibinfo{person}{Daisy~Yi Ding}, \bibinfo{person}{Mykel Kochenderfer}, {and}
  \bibinfo{person}{Jure Leskovec}.} \bibinfo{year}{2020}\natexlab{}.
\newblock \showarticletitle{Handling missing data with graph representation
  learning}.
\newblock \bibinfo{journal}{\emph{arXiv preprint arXiv:2010.16418}}
  (\bibinfo{year}{2020}).
\newblock


\bibitem[\protect\citeauthoryear{Zhang, Maroufy, Chen, and Wu}{Zhang
  et~al\mbox{.}}{2020}]%
        {zhang2020missing}
\bibfield{author}{\bibinfo{person}{Chenguang Zhang}, \bibinfo{person}{Vahed
  Maroufy}, \bibinfo{person}{Baojiang Chen}, {and} \bibinfo{person}{Hulin Wu}.}
  \bibinfo{year}{2020}\natexlab{}.
\newblock \showarticletitle{Missing Data Issues in EHR}.
\newblock \bibinfo{journal}{\emph{Statistics and Machine Learning Methods for
  EHR Data: From Data Extraction to Data Analytics}} (\bibinfo{year}{2020}),
  \bibinfo{pages}{149}.
\newblock


\bibitem[\protect\citeauthoryear{Zhang, Wu, Daigle, Cohen, and Petzold}{Zhang
  et~al\mbox{.}}{2016}]%
        {zhang2016identification}
\bibfield{author}{\bibinfo{person}{Yuanyang Zhang}, \bibinfo{person}{Tie~Bo
  Wu}, \bibinfo{person}{Bernie~J Daigle}, \bibinfo{person}{Mitchell Cohen},
  {and} \bibinfo{person}{Linda Petzold}.} \bibinfo{year}{2016}\natexlab{}.
\newblock \showarticletitle{Identification of disease states associated with
  coagulopathy in trauma}.
\newblock \bibinfo{journal}{\emph{BMC medical informatics and decision making}}
  \bibinfo{volume}{16}, \bibinfo{number}{1} (\bibinfo{year}{2016}),
  \bibinfo{pages}{1--9}.
\newblock


\end{thebibliography}

\begin{appendices}

\section{UCSF MOF data statistics}
For numerical variables, all features except age are list as: feature name, unit: type, description, mean (standard deviation) of no MOF patients, mean (standard deviation) of MOF patients. For age, we show the min-max age for no MOF and MOF patients.\\
For categorical variable, all features are list as: feature name, unit: type, description, number and percentage in each level. \\
\textbf{Demographic:}\\
Gender, no.($\%$): categorical, Male = 1, Female = 0, 1608 (81.3\%), 157 (86.3\%)\\
Age, year: numerical, age of patients, 37(15.0 - 100.0), 45.5 (19.9 - 99.0)\\
BMI, kg/m2: numerical, body mass index, 26.50 $\pm$ 4.80, 27.02 $\pm$ 5.25\\
TBI, no.($\%$): categorical, traumatic brain injury, Yes = 1, No = 0, 629 (31.8\%), 129(70.9\%)\\
\textbf{Injury measurement}\\
AIS-Head: numerical, abbreviated injury scale: head, 1.52 $\pm$ 2.00, 3.40 $\pm$ 4.97\\
AIS-Chest: numerical, abbreviated injury scale: chest, 0.98  $\pm$ 1.53, 2.03 $\pm$ 1.77\\
ISS: numerical, injury severity score, 14.95 $\pm$ 14.46, 33.03 $\pm$ 13.98\\
GCS: numerical, GCS (Glasgow Coma Scale), 9.58 $\pm$ 5.36, 4.92 $\pm$ 3.67\\
\textbf{Admission day} \\
Vasopressor, no.($\%$): categorical, vasopressor utilization Yes = 1, No = 0, 429 (21.7\%), 109 (60.0\%)\\
Heparin\_gtt, no.($\%$): categorical, heparin utilization Yes = 1, No = 0, 84 (4.2\%), 43 (23.6\%)\\
Factor VII treatment, no.($\%$): categorical, factor VII medication given Yes = 1, No = 0, 12 (0.6$\%$), 14 (7.7$\%$)\\
Thromboembolic complication, no.($\%$): categorical, thromboembolic complication condition Yes = 1, No = 0, 83 (4.2\%), 51 (28.0\%)\\
numribfxs: numerical, number of rib fractures, 0.68 $\pm$ 1.97, 3.40 $\pm$ 4.97\\
 \textbf{Initial hour measurement} \\
 WBC,$10^3$/mcL: numerical, white blood cell, 10.44 $\pm$ 4.76, 11.85 $\pm$  5.47\\
HCT, $\%$: numerical, hematocrit, 40.99 $\pm$ 5.48, 39.69 $\pm$ 6.03\\
HGB, g/dL: numerical, hemoglobin, 13.66 $\pm$ 1.74, 13.18 $\pm$ 2.12\\
Bun, mg/dL: numerical, blood urea nitrogen, 15.58 $\pm$ 8.06, 19.66 $\pm$ 15.25\\
Creatinine, g/24 hr: numerical, creatinine value, 1.00 $\pm$ 0.48, 1.27 $\pm$ 1.29\\
D-Dimer, mg/L: numerical, D-Dimer value, 2.90 $\pm$ 5.70, 6.43 $\pm$ 9.13\\
Factor VII, $\%$: numerical, factor VII value, 79.39 $\pm$ 37.82, 75.81$\pm$ 32.57\\
PLTs, $10^3$/ mcL: numerical, platelets, 272.08 $\pm$ 85.04, 267.12 $\pm$ 87.71\\
PTT, sec: numerical, partial thromboplastin time, 29.66 $\pm$ 11.20, 11.08 $\pm$ 0.28\\
Serumco2, mmol/L: numerical, carbon dioxide, 23.66 $\pm$ 4.35, 4.83 $\pm$ 0.28\\
 \textbf{Vital signs} \\
HR, beats per minute: numerical, heart rate, 96.44 $\pm$ 27.74, 102.14 $\pm$ 29.27\\
Respiratory, breaths per minute: numerical, respiratory rate, 19.67 $\pm$ 5.43, 20.57 $\pm$ 5.96\\
SBP, mmHg: numerical, systolic blood pressure, 136.53 $\pm$ 31.22, 132.57 $\pm$ 37.14\\
\textbf{ICU first day measurement}\\
Blood\_units, unit: numerical, blood units transfusion, 2.41 $\pm$ 6.51,9.28 $\pm$ 15.65\\
Crystalloids, ml: numerical, crystalloids for fluids resuscitation, 3856.62 $\pm$ 3395.06, 6721.5 $\pm$ 4437.48\\
FFP\_Units, unit: numerical, fresh frozen plasma, 1.52 $\pm$ 4.71, 7.53 $\pm$ 12.91\\

\section{hyperparameters}
To support the reproducibility of the results in this paper, we provide the hyperparameters we used in all the experiments.
\subsection{PhysioNet sepsis dataset:}
\textbf{Model:} \\
\textbf{Simple imputation \& Classifier}:
epochs: 30, batch size: 128, learning rate: 0.0005-0.002, classifier's weight decay: 5e-4.\\
\textbf{MICE \& Classifier:}
initial strategy: mean, 
maximum number of imputation iteration: 100, 
tolerance: 0.001.\\
\textbf{GAIN \& Classifier:}
epochs for GAIN: 20, batch size for GAIN: 128,  generator's learning rate: 0.0005-0.002, discriminator's learning rate: 0.0005-0.002, generator's weight decay: 5e-4, discriminator's weight decay: 5e-4, p\_hint: 0.9, alpha: 1, epochs for classifier: 30, batch size for classifier: 128, classifier's learning rate: 0.0005-0.002, classifier's weight decay: 5e-4
\\
\textbf{Classifier-GAIN:}
epochs: 50, batch size: 128, generator's learning rate: 0.0005-0.002, discriminator 's learning rate: 0.0005-0.002, classifier's learning rate: 0.0005-0.002, p\_hint: 0.5, alpha: 20, beta:1, generator's weight decay: 5e-4, discriminator's weight decay: 5e-4, classifier's weight decay: 5e-4.

\subsection{UCSF MOF dataset:}
\textbf{Model:} \\
\textbf{Simple imputation \& Classifier:} epochs: 30, batch size: 16, learning rate: 0.0005-0.002, classifier's weight decay: 5e-4.\\
\textbf{MICE \& Classifier:} initial strategy: mean, 
maximum number of imputation iteration: 100, 
tolerance: 0.001.\\
\textbf{GAIN \& Classifier:} epochs for GAIN: 50, batch size for GAIN: 16, generator's learning rate: 0.0005-0.002, discriminator's learning rate: 0.0005-0.002, generator's weight decay: 5e-4, discriminator's weight decay: 5e-4, p\_hint: 0.9, alpha: 5, epochs for classifier: 30, batch size for classifier: 16, classifier's learning rate: 0.0005-0.002, classifier's weight decay: 5e-4.\\
\textbf{Classifier-GAIN:} epochs: 50, batch size: 128, generator's learning rate: 0.0005-0.002, discriminator's learning rate: 0.0005-0.002, classifier's learning rate: 0.0005-0.002, p\_hint: 0.9, alpha: 5, beta: 1, generator's weight decay: 5e-4, discriminator's weight decay: 5e-4, classifier's weight decay: 5e-4.

\end{appendices}

\end{document}